\documentclass[lettersize,journal]{IEEEtran}
\usepackage{amsmath,amsfonts}
\usepackage{algorithmic}
\usepackage{algorithm}
\usepackage{array}
\usepackage[caption=false,font=normalsize,labelfont=sf,textfont=sf]{subfig}
\usepackage{textcomp}
\usepackage{stfloats}
\usepackage{url}
\usepackage{verbatim}
\usepackage{graphicx}
\usepackage{cite}
\usepackage{booktabs}   
\usepackage{multirow}   
\usepackage{pifont}
\usepackage{tabularx} 
\usepackage{mathrsfs}
\usepackage{booktabs}
\usepackage{multirow}
\usepackage[table]{xcolor}
\newcolumntype{Y}{>{\centering\arraybackslash}X} 

\begin{document}

\title{FCUS-rPPG: A Fast-Converging Unsupervised Framework for Remote Photoplethysmography via Gradient Oscillation Suppression}

\author{Jiajie Li, Yu Liu \IEEEmembership{Senior Member, IEEE}, Rencheng Song \IEEEmembership{Senior Member, IEEE}, Xun Chen \IEEEmembership{Fellow, IEEE}, Juan Cheng \IEEEmembership{Senior Member, IEEE}
\thanks{J. Li, J. Cheng, Y. Liu, R. Song are with the Department of Biomedical Engineering, and also with Anhui Province Key Laboratory of Measuring Theory and Precision Instrument, Hefei University of Technology, Hefei 230009, China (e-mail: lijiajie@mail.hfut.edu.cn; chengjuan@hfut.edu.cn; yuliu@hfut.edu.cn; rcsong@hfut.edu.cn).}
\thanks{Xun Chen is with the Department of Electronic Engineering and Information Science, University of Science and Technology of China, Hefei 230026, China (e-mail: xunchen@ustc.edu.cn).}
\thanks{\textit{(Corresponding author: Juan Cheng.)}}
}

\maketitle

\begin{abstract}
Remote photoplethysmography (rPPG) enables non-contact extraction of blood volume pulse (BVP) signals using consumer-grade cameras.
Recent unsupervised rPPG methods learn BVP representations without requiring ground-truth physiological annotations, yet their optimization is often hindered by noisy and unstable gradients, resulting in slow convergence and limited cross-domain generalization.
In this paper, we propose FCUS-rPPG, a fast-converging unsupervised rPPG framework with strong generalization capability. 
Motivated by the observation that BVP representations exhibit both multi-spectral covariation and low-dimensional manifold structure, we design a spectrally shared backbone that facilitates BVP feature disentanglement while improving optimization efficiency.
To jointly enhance convergence stability and generalization performance, we further develop a unified optimization framework operating at the gradient, loss-landscape, and feature-representation levels.
Specifically, a post-verification masking mechanism filters out misleading gradients according to the weak-amplitude physiological prior of BVP signals; a perturbation-based loss landscape smoothing strategy steers optimization toward more generalizable flat minima; and a noise-aware null-space regularization constrains feature updates to the orthogonal complement of the noise subspace, thereby mitigating noise-induced representation drift.
Extensive experiments on five datasets demonstrate that FCUS-rPPG requires only one training epoch, whereas existing methods typically require tens to hundreds of epochs.
Notably, FCUS-rPPG consistently achieves state-of-the-art (SOTA) performance in cross-dataset evaluations.
This study provides an efficient and robust solution to the real-world deployment of unsupervised rPPG. The source code will be publicly available at https://github.com/JiaJieLee/FCUS-rPPG.

\end{abstract}


\section{Introduction}
\IEEEPARstart{C}{ardiovascular} physiological signals, such as heart rate (HR) and heart rate variability (HRV), provide important indicators of human physical and psychological health.
Although conventional monitoring devices, such as electrocardiography (ECG) \cite{Siontis2021AIECG, Lin2024AIECGAlert} and photoplethysmography (PPG) \cite{Quer2021WearableCOVID, dunn2021wearable}, are widely employed to assess these parameters, their contact-based nature limits long-term and unobtrusive monitoring scenarios \cite{stingeni2015role,eem2024li}.
As a promising alternative, remote photoplethysmography (rPPG) enables non-contact cardiovascular monitoring by recovering blood volume pulse (BVP) signals from subtle facial skin color variations captured by commodity-grade cameras \cite{verkruysse2008remote}. This capability has facilitated emerging applications in telemedicine \cite{mcduff2023, rppgra2019chen}, face anti-spoofing \cite{Liu2018CVPR}, and driver monitoring \cite{10818975}.
However, high-quality BVP recovery is hampered because cardiac-induced facial color variations are extremely subtle and easily corrupted by head motions and illumination variations.

Traditional rPPG methods, built upon the dichromatic reflection model \cite{tominaga1994dichromatic}, leverage chrominance projection \cite{de2013robust, wang2016algorithmic} and blind source separation (BSS) \cite{poh2010non} to suppress noise and recover BVP signals.
Nevertheless, their performance is fundamentally constrained by rigid hand-crafted priors.

\begin{figure}[t]
\centering
\includegraphics[width=0.98\linewidth]{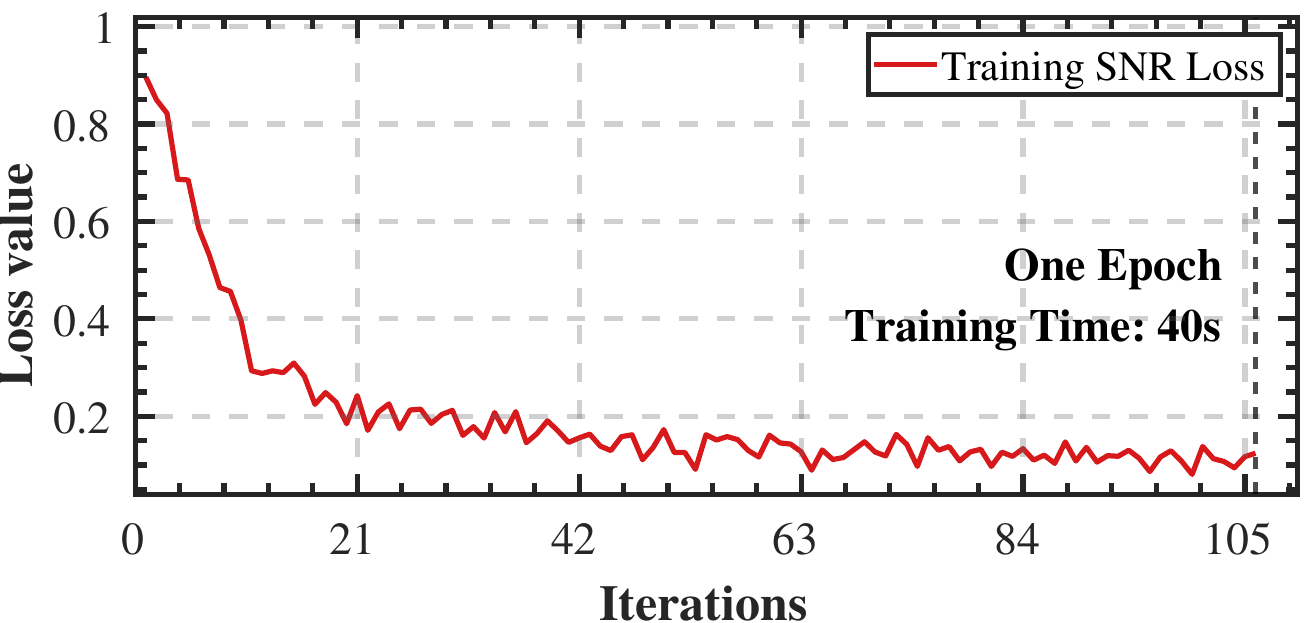}
\caption{SNR loss curves of the raw signals during training from scratch on the UBFC-rPPG dataset. The loss decreases rapidly and converges within a single epoch (106 iterations), requiring only 40 seconds of training time.}
\label{lossubfc}
\end{figure}

Recent supervised deep learning (DL)-based rPPG methods \cite{chen2018deepphys, yu2019remote, niu2019rhythmnet, song2021pulsegan, yu2022physformer} have substantially advanced non-contact BVP estimation.
By leveraging ground-truth (GT) signals to guide the extraction of physiological features during training, these methods achieve impressive BVP recovery performance.
However, acquiring high-quality GT signals necessitates synchronized recording with medical-grade contact sensors, making large-scale real-world data collection costly and cumbersome.

To eliminate the reliance on GT labels, label-free DL-based rPPG methods have recently attracted increasing attention.
Existing studies mainly follow two paradigms: self-supervised representation learning based on pretext tasks \cite{liu2024rppgmae, gideon2021way, sun2022contrast, wang2022slfrpm, yang2022simper, yue2023facial, sun2024contrastplus, MA2026113501}, and unsupervised learning guided by frequency-domain physiological priors \cite{speth2023non}.
Although these methods alleviate the exhaustive annotation burden and enable BVP recovery without supervised fine-tuning, they suffer from slow convergence and compromised generalization due to gradient oscillations.
In practice, they often require tens \cite{sun2024contrastplus, yang2022simper} to hundreds \cite{gideon2021way, liu2024rppgmae, yue2023facial, speth2023non} of training epochs, resulting in substantial computational and time overhead.
Moreover, their performance degrades noticeably under cross-dataset domain shifts \cite{chu2025remember}, limiting real-world generalization capability.

Some studies have explored gradient-aware training strategies to improve generalization in unsupervised rPPG, such as synthetic gradient prediction \cite{meta2020lee} and gradient modification techniques \cite{chu2025remember}.
However, these methods are not designed for rapid convergence from scratch and still rely on computationally expensive pre-training for parameter initialization.
Rapid convergence to generalizable minima remains challenging for unsupervised rPPG methods trained from scratch.

From a gradient-based optimization perspective, the slow convergence and poor generalization of unsupervised rPPG can be jointly attributed to four factors.
\textbf{First}, optimization barriers in representation learning.
The extremely weak amplitude of skin BVP signals, coupled with substantial inter-subject heterogeneity \cite{verkruysse2008remote}, hinders the optimization of consistent physiological representations.
\textbf{Second}, gradient contamination from misleading samples.
Since existing unsupervised rPPG methods heavily rely on frequency-domain priors \cite{gideon2021way, sun2022contrast, wang2022slfrpm, yang2022simper, speth2023non, yue2023facial, sun2024contrastplus, liu2024rppgmae}, in-band non-physiological periodic components can easily bias optimization toward spurious convergence directions.
\textbf{Third}, rugged loss landscapes with high local variance.
Limited mini-batch sizes and the absence of strong GT supervision introduce substantial local gradient variance, rendering stochastic gradient descent (SGD) unstable and prone to converging to poorly generalizing sharp minima \cite{Cha2021}.
\textbf{Fourth}, trajectory distortion induced by complex noise.
Due to the inherently low signal-to-noise ratio (SNR) of rPPG and the non-stationary properties of practical noise \cite{deHaan2014}, noise-induced gradients can overwhelm valid physiological optimization directions, severely destabilizing training dynamics.

To address the above challenges, we propose a \textbf{F}ast-\textbf{C}onverging \textbf{U}n\textbf{S}upervised \textbf{rPPG} framework, termed \textbf{FCUS-rPPG}, which achieves rapid convergence while maintaining strong generalization capability.
Specifically, we revisit the rPPG problem from a novel perspective that unifies multi-spectral physiological covariation with the low-dimensional manifold structure of BVP representations.
Guided by this insight, we develop a Low-dimensional Spectrally-Shared (LSS) backbone.
By enforcing spectrally shared and lightweight parameterization, LSS projects rPPG signals onto a compact physiological manifold, thereby reducing optimization complexity while extracting domain-invariant physiological representations.
To block contaminated gradient backpropagation, we further introduce an amplitude-prior-guided Post-verification Gradient Masking (PGM) mechanism.
By reformulating the weak-amplitude property of BVP signals as a physiological prior, PGM performs physiological plausibility validation on model outputs to derive gradient masks, thereby preventing spurious convergence while preserving data diversity.
To steer SGD toward more generalizable flat minima \cite{Cha2021}, we propose a Loss Landscape Smoothing (LLS) strategy.
By estimating the expected loss under parallel data perturbations, LLS suppresses high-variance ruggedness in the optimization space, thereby promoting convergence toward robust flat optima.
Finally, to mitigate noise-induced trajectory distortion, we introduce a Noise-aware Null-space Regularization (NNR) strategy.
NNR estimates noise representations through a noise-aware branch equipped with zero-phase band-stop filtering and noise reconstruction task.
The learned BVP representations are subsequently constrained to evolve within the orthogonal complement of the noise subspace, thereby improving robustness against noise-driven gradient interference.

In summary, our contributions are four-fold:
\begin{enumerate}

\item We propose FCUS-rPPG, an unsupervised rPPG framework that achieves rapid convergence from scratch within a single training epoch, substantially reducing the optimization cost compared with existing methods that typically require tens to hundreds of epochs, while preserving strong cross-dataset generalization capability. To the best of our knowledge, this is the first work to study  convergence acceleration in unsupervised rPPG.

\item We revisit the rPPG problem from the novel perspective of the multi-spectral covariation and the low-dimensional manifold structure underlying BVP representations. Motivated by this insight, we develop a low-dimensional spectrally shared backbone that reduces optimization complexity while enabling robust extraction of domain-invariant physiological features.

\item We develop a unified optimization framework to jointly improve convergence efficiency and gradient reliability, comprising: i) a post-verification gradient masking mechanism that filters out contaminated gradient updates; ii) a perturbation-driven loss landscape smoothing strategy that guides optimization toward more generalizable flat minima; and iii) a noise-aware null-space regularization module that mitigates noise-induced gradient interference during training.

\item Extensive experiments conducted on five datasets demonstrate the convergence efficiency and generalization capability of FCUS-rPPG. Remarkably, even when trained from scratch for only a single epoch (as exemplified in Fig. \ref{lossubfc}), the proposed method consistently outperforms existing state-of-the-art (SOTA) approaches under challenging cross-dataset evaluation settings.

\end{enumerate}

\begin{figure*}[t]
\centering
\includegraphics[width=1\linewidth]{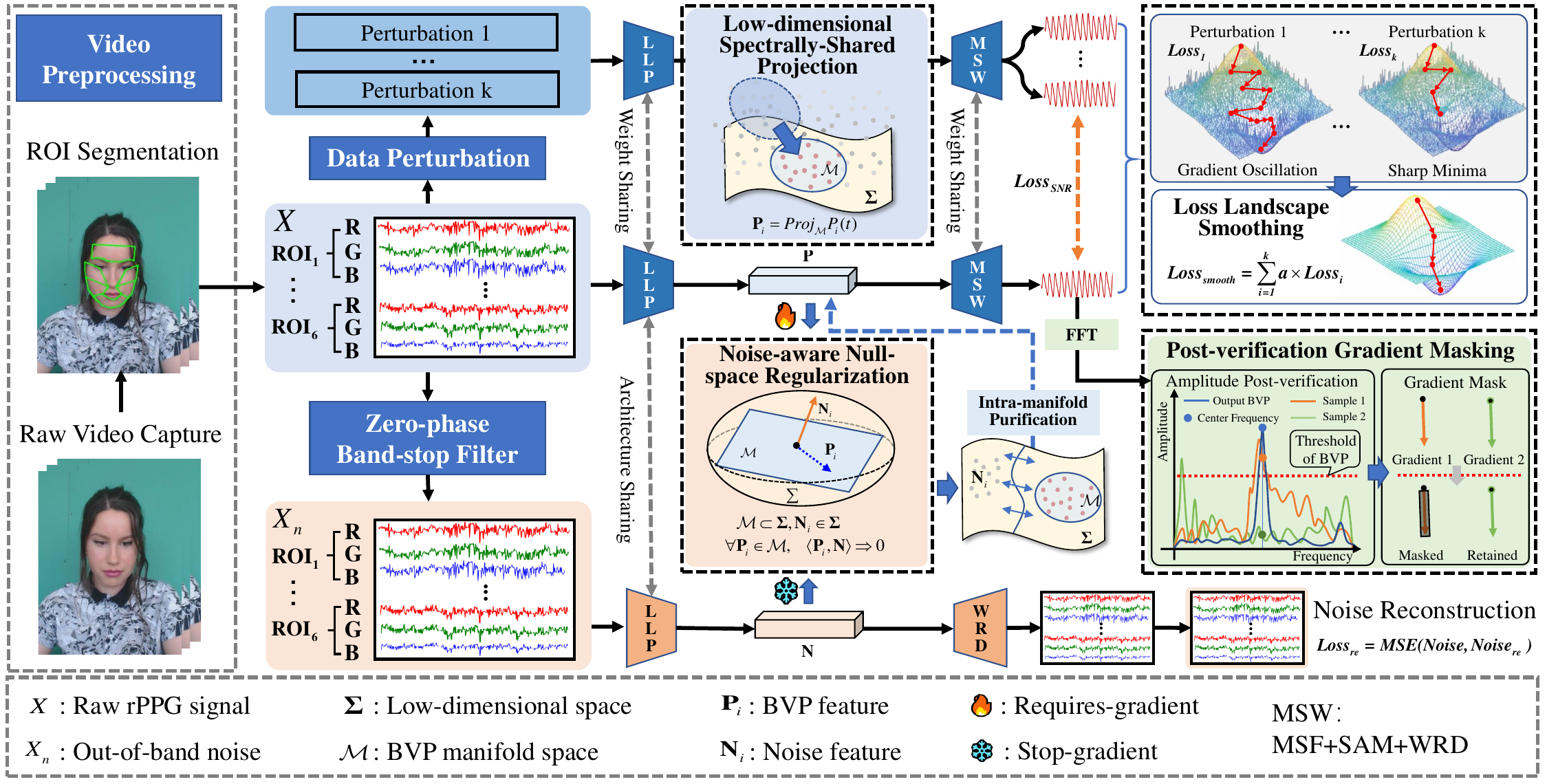}
\caption{The overall pipeline and key components of the proposed FCUS-rPPG framework. The framework comprises video preprocessing, a low-dimensional spectrally-shared backbone (LLP+MSW), and a unified optimization acceleration framework. Notably, the unified optimization framework achieves synergistic constraints via post-verification gradient masking, loss landscape smoothing, and noise-aware null-space regularization.}
\label{framework}
\end{figure*}
\section{Related Work}
\label{sec:relatedwork}

\subsection{Remote Physiological Measurement}
Verkruysse et al. \cite{verkruysse2008remote} pioneered rPPG for facial BVP extraction using consumer cameras under ambient light, revealing that while all camera-captured red, green, and blue (RGB) spectral channels contain pulsatile information, wavelength-dependent blood light absorbance and skin penetration depths yield the strongest BVP signal in the green channel.
Subsequently, Poh et al. \cite{poh2010non} formulated the RGB channel signals as a linear mixture of the BVP and other independent noise sources, successfully extracting the BVP signal by introducing BSS techniques.
To further suppress interferences from ambient illumination variations and motion artifacts, De Haan et al. \cite{de2013robust} and Wang et al. \cite{wang2016algorithmic} proposed orthogonal projection in the RGB color space based on skin optical reflection models, thereby extracting BVP signals.

In recent years, supervised DL-based rPPG methods \cite{chen2018deepphys, yu2019remote, liu2020multi, song2021pulsegan, yu2022physformer, Lu2023NEST, liu2023RRPEtip, zhang2025agrtip} have achieved superior performance in BVP recovery.
Supervised by GT physiological signals, these methods leverage deep neural networks to extract facial BVP features, typically utilizing shallow-layer spectral-mixing modeling to directly map raw RGB signals into high-dimensional representations.
Specifically, supervised DL methods for rPPG fall into two main paradigms: end-to-end and hand-crafted feature-based approaches.
End-to-end methods employ 2D CNNs \cite{chen2018deepphys}, 3D CNNs \cite{yu2019remote}, or Transformers \cite{yu2022physformer} to directly extract spatiotemporal physiological features from raw facial videos.
Conversely, hand-crafted feature-based methods \cite{niu2019rhythmnet, Lu2023NEST, zhang2025agrtip} simplify the learning task by extracting facial regions of interest (ROIs) and transforming them into intermediate representations, such as temporal signals or spatiotemporal maps (STMaps), prior to network mapping.
Regardless of the chosen paradigm, supervised methods rely on large-scale, high-quality GT labels, the acquisition and strict alignment of which incur prohibitive time and labor costs.
To break through this bottleneck, researchers have gradually shifted their focus towards label-free DL paradigms.

\subsection{Label-Free Deep Learning for rPPG}
Label-free DL-based rPPG methods, which eliminate the need for GT label supervision during training and can even achieve BVP recovery without supervised fine-tuning, can be primarily classified into two main branches: self-supervised learning (SSL) and unsupervised learning.

Specifically, SSL seeks to extract the intrinsic representations of unlabeled data through meticulously designed pretext tasks, such as contrastive learning \cite{lecun2006, Wu2018CVPR, oord2018representation} and masked reconstruction \cite{He2022CVPR}.
Gideon et al. \cite{gideon2021way} first introduced contrastive learning to this field by constructing positive and negative sample pairs of video clips through a frequency resampling strategy.
Subsequent studies further optimized the pair construction mechanism based on the frequency-domain characteristics of physiological signals \cite{gideon2021way, sun2022contrast, wang2022slfrpm, yang2022simper, yue2023facial, sun2024contrastplus}.
In terms of masked reconstruction, Liu et al. \cite{liu2024rppgmae} employed a masked autoencoder (MAE) framework to reconstruct masked STMaps, enabling BVP recovery with minimal labeled data fine-tuning after self-supervised pre-training.
Ma et al. \cite{MA2026113501} synergized contrastive learning and MAE to achieve SSL based on unlabeled STMaps.
Recent advances \cite{speth2023non, li2026udarppg} have highlighted the significant potential of unsupervised learning for rPPG. In particular, Speth et al. \cite{speth2023non} proposed the first unsupervised learning-based rPPG framework, which successfully exploits intrinsic periodic skin-color variations from unlabeled facial videos through the incorporation of frequency-domain priors and cumulative distribution function constraints.

Despite the significant advancements achieved by these label-free methods, they generally suffer from an excessive number of training epochs.
Specifically, the original studies report that \cite{liu2024rppgmae} necessitates 400 pre-training epochs and 30 fine-tuning epochs, \cite{speth2023non} requires 200 epochs, \cite{gideon2021way,yue2023facial} requires 100 epochs, and \cite{MA2026113501} involves 150 pre-training epochs and 45 fine-tuning epochs. Even the more efficient \cite{sun2024contrastplus} still requires 30 epochs.
This iterative training paradigm over large-scale data inevitably incurs substantial computational overhead and time costs.
More critically, the learned parameters are susceptible to severe overfitting, falling into sharp minima \cite{Cha2021} that compromise generalization.
To overcome these limitations, we propose a novel unsupervised framework, FCUS-rPPG, which achieves convergence within a single epoch on identical datasets while maintaining superior generalization performance.

\section{Methodology}
\label{sec:method}
This section introduces the overall pipeline of the proposed FCUS-rPPG framework, as shown in Fig. 2. Motivated by the theoretical analysis of BVP features from the perspectives of multi-spectral physiological covariation and low-dimensional physiological manifolds, we design a low-dimensional spectrally shared backbone to achieve domain-invariant BVP feature projection. To accelerate optimization while preserving generalization, we further introduce a unified optimization acceleration framework that enforces synergistic constraints at three levels, namely gradients, loss landscape, and feature representations, through post-verification gradient masking, loss landscape smoothing, and noise-aware null-space regularization, respectively. The subsequent sections detail each core component.

\subsection{Theoretical Formulation}
\label{problem}
\begin{figure}[t]
\centering
\includegraphics[width=1\linewidth]{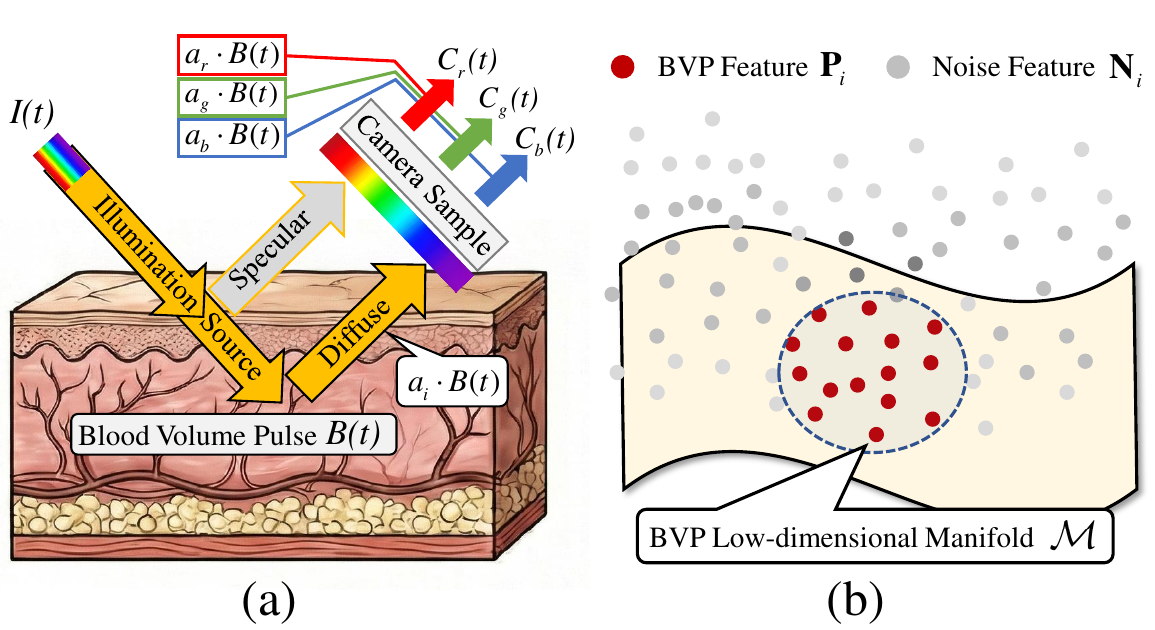}
\caption{(a) Multi-spectral physiological covariation and (b) low-dimensional physiological manifold hypothesis.}
\label{theore}
\end{figure}

\subsubsection{Multi-Spectral Physiological Covariation}
\label{mpc}
Based on the dichromatic reflection model \cite{tominaga1994dichromatic}, facial video signals captured under ambient illumination are composed of specular and diffuse reflections.
The diffuse reflection component encompasses not only the static skin-tissue color but also the physiological information induced by variations in subcutaneous BVP.
As shown in Fig. \ref{theore} (a), by independently observing multi-spectral information via camera-based sampling, the discrete spectral channels provide synchronized and multidimensional measurements of the same BVP process.

For spectral channel $i$, the corresponding time-domain observed signal, $C_i(t)$, can be expressed as follows:
\begin{equation}
C_i(t) = I_i(t) \cdot [(a_i B(t) + D_i) + S_i(t)] + n_i(t)
\end{equation}
where $B(t)$ denotes the skin BVP;
$a_i$ defines the blood absorption coefficient associated with spectral channel $i$; $I_i(t)$ indicates the luminance intensity level;
$D_i$ denotes the static skin-tissue diffuse reflection;
$S_i(t)$ signifies the specular reflection;
and $n_i(t)$ accounts for the camera sensor noise.
Eq. (1) can be further decoupled into a target pulsatile component, $P_i(t)$, and a generalized noise component, $N_i(t)$, which encompasses the skin-tissue direct current component:
\begin{equation}
C_i(t) =
\underbrace{I_i(t)(a_i \cdot B(t))}_{P_i(t)}
+
\underbrace{I_i(t)(D_i + S_i(t)) + n_i(t)}_{N_i(t)}
\end{equation}
Although standard cameras typically capture RGB signals, Eq. (2) extends naturally to broader spectral bands, such as the near-infrared (NIR).

The signal intensity of the pulsatile component $P_i(t)$ exhibits a two-fold variability. Cross-spectrally, the captured intensity diverges across channels due to modulation by parameters $I_i(t)$ and $a_i$. Within a single spectrum, this intensity further fluctuates depending on the individual subject, the illumination spectral distribution, and the acquisition device.
Even though these effects are intertwined in real-world scenarios, all spectral channels share a unified optimization objective: to recover the underlying weak, quasi-periodic BVP source signal $B(t)$.

\subsubsection{Low-Dimensional Physiological Manifold Hypothesis}
\label{lpmh}
To further streamline the unified optimization objective and mitigate the impact of $\mathbf{N}_i(t)$, we propose the low-dimensional physiological manifold hypothesis, as illustrated in Fig. \ref{theore} (b).

Specifically, a mapping $\Phi : \mathbb{R}^W \rightarrow \mathbb{R}^{D}$ is introduced to project the $W$-dimensional temporal window observation $\mathbf{C}_i(t)$ into an information-complete feature space. Within this $D$-dimensional space, the observation window $\mathbf{C}_i(t)$ is mapped to a feature vector $\mathbf{X}_i^{\Phi}(t)$, which is assumed to approximately admit the following additive decomposition:
\begin{equation}
    \mathbf{X}_i^{\Phi}(t) = \Phi(\mathbf{C}_i(t)) = \mathbf{P}_i^{\Phi}(t) + \mathbf{N}_i^{\Phi}(t),
\end{equation}
where $\mathbf{P}_i^{\Phi}(t)$ and $\mathbf{N}_i^{\Phi}(t)$ represent the physiological and generalized noise projections, respectively.

Owing to the quasi-periodicity of the cardiac cycle, the high-dimensional physiological feature $\mathbf{P}_i^{\Phi}(t)$ is constrained to a compact low-dimensional manifold $\mathcal{M} \subset \mathbb{R}^D$. This manifold is parameterized by a mapping $g: \mathbb{R}^d \rightarrow \mathbb{R}^D$:
\begin{equation}
    \mathcal{M} = \{ g(\mathbf{z}) \mid \mathbf{z} \in \mathbb{R}^{d} \},
\end{equation}
where $\mathbf{z}$ represents a latent variable encoding the intrinsic physiological state. The intrinsic dimensionality $d$ of the manifold is strictly smaller than the ambient feature dimension $D$ (i.e., $d \ll D$), inherently ensuring that $\mathbf{P}_i^{\Phi}(t) \in \mathcal{M}$.

Unlike the compact physiological manifold, the noise representation $\mathbf{N}_i^{\Phi}(t)$ diffuses broadly across the feature space due to its complex stochastic variations, implying a non-singular covariance matrix $\text{Cov}(\mathbf{N}_i^{\Phi})$ with $\text{rank}(\text{Cov}(\mathbf{N}_i^{\Phi})) = D$. This geometric separability between the low-dimensional manifold and high-dimensional noise dispersion provides theoretical guidance for decoupling the BVP signal from complex noise.

To further simplify the extraction process, although the physiological manifold $\mathcal{M}$ may exhibit a globally nonlinear topology, we assume it admits a robust linear approximation within a localized temporal context. Consequently, we introduce a linear operator $\mathcal{P}$ to directly project the original observation signal $\mathbf{C}_i(t)$ into the $d$-dimensional linear subspace $\mathcal{V}^{d}$ that locally approximates $\mathcal{M}$:
\begin{equation}
\mathbf{z}_i(t) = \mathcal{P} \{ \mathbf{C}_i(t) \}, \quad \text{s.t.} \;\; \mathbf{z}_i(t) \in \mathcal{V}^{d},
\end{equation}
where $\mathbf{z}_i(t) \in \mathbb{R}^{d}$ denotes the resulting low-dimensional physiological representation.

\subsection{Video Preprocessing}
\label{DP}
We leverage the MediaPipe \cite{lugaresi2019mediapipe} algorithm to extract facial landmarks from video frames and delineate the facial ROIs following  \cite{niu2020video}.
For the $t$-th frame, we acquire the raw signals of each ROI via spatial pixel averaging and concatenate them into a spatiotemporal tensor to formulate the raw rPPG signal $X \in \mathbb{R}^{R \times C \times T}$, where $R$, $C$, and $T$ denote the number of ROIs, spectral channels, and video frames, respectively.
Most hand-crafted methods \cite{niu2020video,niu2019rhythmnet, Lu2023NEST, zhang2025agrtip} employ sliding-window normalization to mitigate skin-tone-induced DC offsets and unify rPPG amplitudes. However, this operation introduces arbitrary scaling factors that severely distort the authentic BVP amplitude. To preserve these physiological properties, we bypass normalization and directly feed $X$ into the DL model.

\subsection{Low-dimensional Spectrally-shared Backbone}
\label{LSS}

\begin{figure}[t]
\centering
\includegraphics[width=1\linewidth]{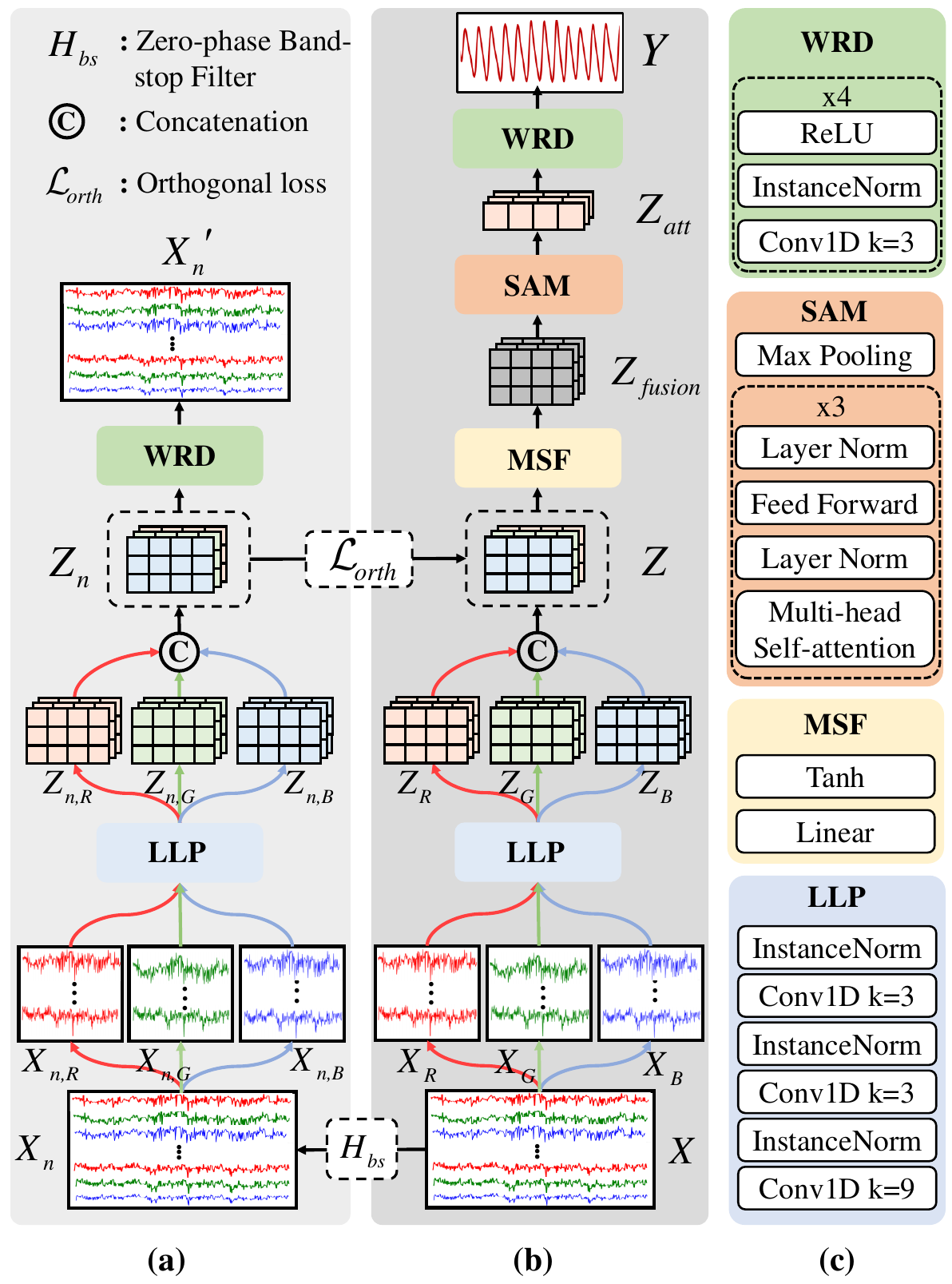}
\caption{Overall pipeline of the proposed Low-dimensional Spectrally-Shared (LSS) backbone and Noise-aware Null-space Regularization (NNR). (a) NNR branch. (b) LSS branch. (c) Detailed network architecture of the proposed model.}
\label{model}
\end{figure}

Motivated by the theoretical formulation introduced in Section \ref{mpc}, we propose the LSS backbone to extract the intrinsic BVP signal from multi-spectral inputs.
As illustrated in Fig. \ref{model}(b), the proposed backbone operates through a systematic pipeline comprising four sequential modules: Low-dimensional Linear Projection (LLP), Multi-Spectral Fusion (MSF), Spatial Attention Modulation (SAM), and a Waveform Reconstruction Decoder (WRD).

Specifically, the LLP module comprises three sequential blocks, each integrating a 1D Convolutional Neural Network (1D-CNN) with Instance Normalization (IN).
Notably, to preclude the early cross-contamination of spatial noise, we enforce strictly ROI-independent temporal modeling during this stage.
Furthermore, IN is utilized in lieu of conventional Batch Normalization (BN) to mitigate batch-size dependency and accelerate network convergence.
Given the input tensor $X$, the initial block extracts low-level temporal features, while the subsequent two blocks progressively abstract the features and project them into a compact $C_l$-dimensional subspace (where $C_l = 16$). Specifically, let $X_i \in \mathbb{R}^{R \times 1 \times T}$ ($i \in {R, G, B}$) denote the single-channel spectral observation. The low-dimensional linear projection, parameterized by $\Theta_{llp}$, is formulated as:
\begin{equation}
Z_i = \mathcal{F}(X_i; \Theta_{llp}),
\label{eq:llp_projection}
\end{equation}
where $\mathcal{F}(\cdot; \Theta)$ represents the parameterized mapping function, and $Z_i \in \mathbb{R}^{R \times C_l \times T}$ represents the derived low-dimensional manifold representation for the $i$-th spectral BVP signal.

Subsequently, the channel-specific representations $Z_i$ are concatenated along the feature dimension to construct a joint spectral representation $Z \in \mathbb{R}^{R \times 3C_l \times T}$. This joint tensor is then processed by the MSF module, consisting of a single linear layer followed by a $\tanh$ activation function, to model cross-spectral correlations. Parameterized by $\Theta_{msf}$, the spectrally fused feature $Z_{fusion} \in \mathbb{R}^{R \times C_l \times T}$ is expressed as:
\begin{equation}
Z_{fusion} = \mathcal{F}(Z; \Theta_{msf}).
\label{eq:msf_fusion}
\end{equation}

Operating on the fused spectral representation $Z_{fusion}$, the SAM module performs adaptive spatial attention to emphasize pulse-bearing regions, producing the spatially aggregated tensor $Z_{att} \in \mathbb{R}^{1 \times C_l \times T}$.
As depicted in Fig. \ref{model}(c), we perform global aggregation of multi-ROI features using three stacked spatial self-attention blocks.
Each block consists of multi-head self-attention, a feed-forward network, and layer normalization.
This attention mechanism enables the network to perform dynamic and adaptive modeling of the spatial dependencies across different ROIs.
Subsequently, a max-pooling operation is applied along the spatial dimension to extract the temporal representation from the most informative ROI.

Finally, the WRD, composed of four interleaved 1D-CNN layers, IN, and ReLU activations, decodes the feature to reconstruct a high-quality BVP waveform, denoted as $Y$.

\subsection{Post-verification Gradient Masking}
\label{PGM}

\begin{figure}[t]
\centering
\includegraphics[width=1\linewidth]{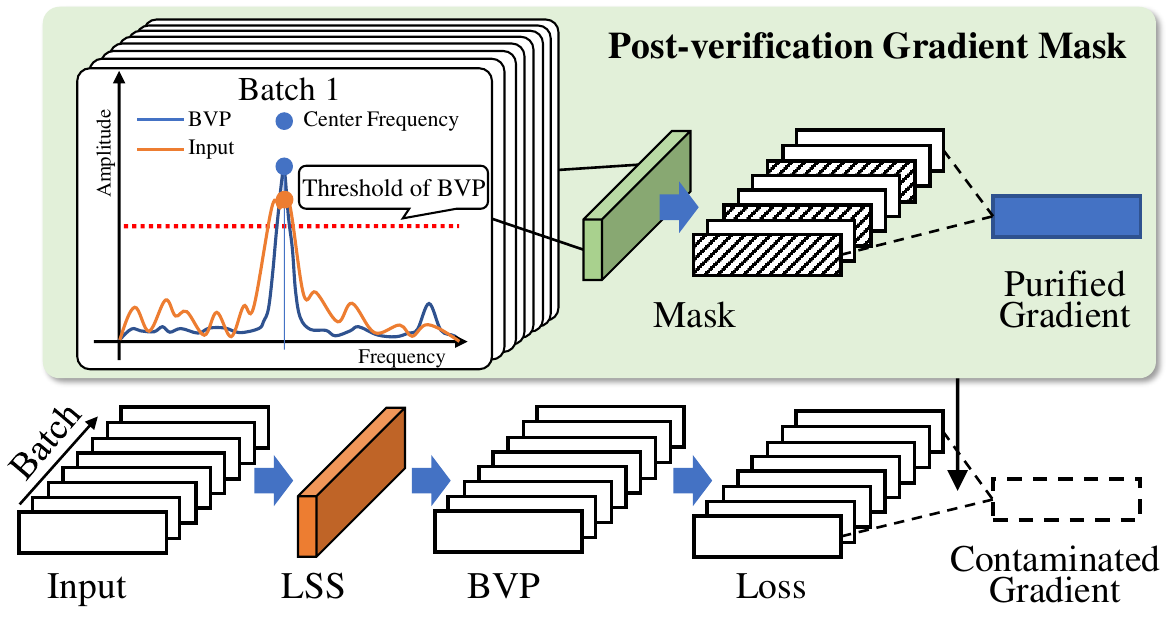}
\caption{Illustration of the Post-verification Gradient Masking (PGM) mechanism.}
\label{MRAM}
\end{figure}

In unsupervised rPPG learning, the frequency-domain distribution of the BVP signal serves as a critical physiological prior.
However, in the absence of GT supervision, periodic noise components that overlap with this predefined frequency band (such as exercise-induced oscillatory motion artifacts or driving-related cyclic illumination variations) can be erroneously extracted as the target BVP signal.
Non-physiological periodic noise receives low penalties, leading to spurious convergence that misguides gradient-based optimization.

While the subtle amplitude of the BVP signal is typically considered a fundamental bottleneck, it uniquely serves as a crucial physiological constraint for unsupervised calibration.
Since non-physiological disturbances generally exhibit substantially higher energy than the target pulse, common motion artifacts and illumination variations become inherently separable in the amplitude domain \cite{Wang2017}.
Nevertheless, imposing amplitude thresholds over the entire physiological band of raw signals would inevitably discard a large number of samples, severely compromising data diversity.
To resolve this dilemma, we propose a post-verification gradient masking mechanism guided by the amplitude prior, which performs verification only on the model-predicted center-frequency component rather than the entire physiological band.

Specifically, let $\Theta_{\text{lss}}$ denote the parameters of the LSS model, and $X_b \in \mathbb{R}^{R \times C \times T}$ be the raw input tensor for the $b$-th sample ($1 \le b \le B$) in a minibatch of size $B$. Here, $X_{b}^{c, r} \in \mathbb{R}^{T}$ represents the raw time-domain signal corresponding to the $c$-th spectral channel and the $r$-th ROI. The intrinsic BVP signal recovered by the LSS backbone is formulated as $Y_b = \mathcal{F}(X_{b}; \Theta_{\text{lss}})$.

To extract the quasi-periodic physiological signals from the raw input $X_b$, we formulate a SNR loss on the output $Y_b$ of the LLS backbone. First, we compute the normalized Power Spectral Density (PSD), denoted as $S_n(f)$, by applying the Fast Fourier Transform (FFT):
\begin{equation}
S_n(f) = \frac{\left| \mathscr{F}\{Y_b\}(f) \right|^2}{\sum_{f} \left| \mathscr{F}\{Y_b\}(f) \right|^2},
\end{equation}
where $\mathscr{F}\{\cdot\}$ denotes the FFT operation, and $\left| \cdot \right|^2$ computes the spectral power. Next, we extract the center
frequency $f_b$ by locating the peak of the amplitude spectrum within the physiological frequency band $\Omega$:
\begin{equation}
f_b = \arg\max_{f \in \Omega} \left| \mathscr{F}\{Y_b\}(f) \right|.
\end{equation}
Based on the normalized PSD and the estimated center frequency $f_b$, the SNR loss $\mathcal{L}_{\text{SNR}}^{(b)}$ is formulated as:
\begin{equation}
\mathcal{L}_{SNR}^{(b)} = 1 - \sum_{f = f_b - \Delta f}^{f_b + \Delta f} S_n(f),
\end{equation}
where $\Delta f$ denotes the permissible range of HR variation within the segmented signal.

To apply the masking mechanism, we evaluate the frequency domain representation of the raw input.
The maximum amplitude across all ROIs and spectral channels at the center frequency $f_b$ is defined as:
\begin{equation}
A_b = \max_{c, r} \left| \mathscr{F}\{X_{b}^{c, r}\}(f_b) \right|.
\label{eq:peak_amp}
\end{equation}
Using this peak amplitude $A_b$, we introduce a predefined threshold $\tau$ to construct an indicator function $\mathbb{I}(\cdot)$. This function generates a binary mask $M_b$:
\begin{equation}
M_b = \mathbb{I}(A_b \le \tau),
\label{eq:binary_mask}
\end{equation}
where $M_b \in \{0, 1\}$.
Finally, the masked loss $\mathcal{L}_{mask}$ is formulated as a normalized weighted sum over the valid sample count, ensuring that this truncation at the loss level naturally translates into the desired gradient masking mechanism during backpropagation, given by:
\begin{equation}
\mathcal{L}_{mask} = \frac{1}{B - m + \epsilon} \sum_{b=1}^{B} M_b \mathcal{L}_{SNR}^{(b)},
\label{eq:masked_loss}
\end{equation}
where $m$ denotes the total number of excluded samples in the current batch and $\epsilon$ is a small constant added for numerical stability to prevent division by zero.

\subsection{Loss Landscape Smoothing}
\label{LLS}
\begin{figure}[t]
\centering
\includegraphics[width=1\linewidth]{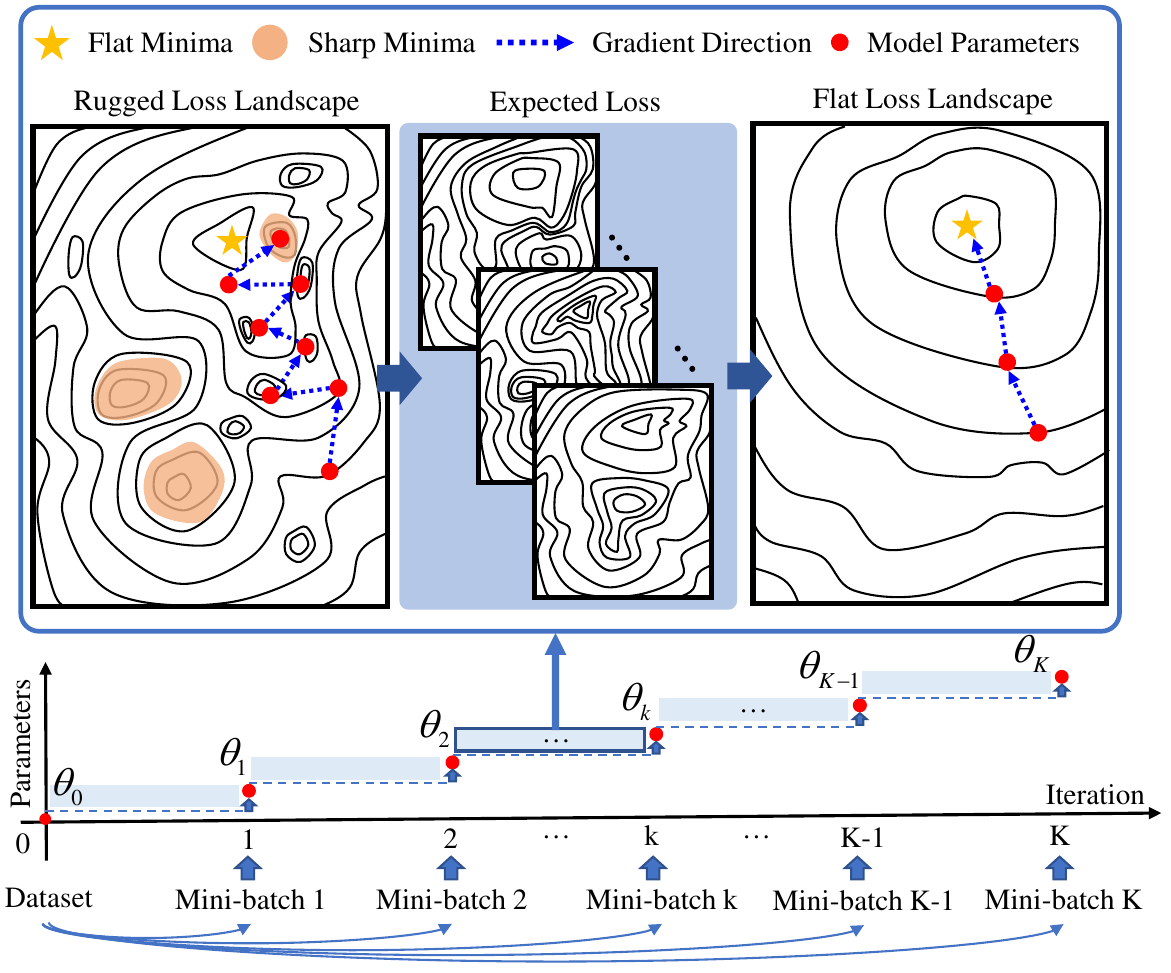}
\caption{Illustration of the Loss Landscape Smoothing (LLS) strategy.}
\label{figLLS}
\end{figure}
The ideal optimization objective is to converge to the global optimum $\theta^*$ on the loss landscape defined over the complete data distribution, thereby theoretically achieving optimal generalization.
However, the complete data distribution is inaccessible, and full-batch training on large-scale datasets is computationally prohibitive. 

Consequently, SGD is widely adopted, where the optimization navigates a sequence of stochastic local loss landscapes induced by mini-batches.
These landscapes can be regarded as perturbed approximations of the global objective, typically exhibiting higher ruggedness and batch-induced sharp minima. As a result, the update trajectory may deviate from the true gradient direction toward $\theta^*$, leading to slower convergence and an increased risk of converging into suboptimal sharp minima, thereby degrading generalization performance.

To address the above issue, this paper proposes a LLS strategy based on expected loss.
As illustrated in Fig. \ref{figLLS}, rather than naively increasing the batch size, the proposed method constructs parallel inputs by applying diverse data perturbations to the raw signal $X$.
By optimizing the expected loss over these augmented variants, the method stabilizes the loss landscape.
It promotes consistent flat minima and mitigates batch-induced sharp ones, thereby improving data utilization per iteration.
Furthermore, owing to the highly lightweight architecture of the LSS backbone, incorporating these parallel computations introduces only negligible computational overhead.

Specifically, we design three parallel data perturbation strategies: (1) skin tone and amplitude alteration, (2) Gaussian noise injection, and (3) local dropout.
The corresponding augmented signals, denoted as $X_s$, $X_g$, and $X_d$, are detailed below.

\subsubsection{Skin Tone and Amplitude Alteration}
By utilizing the tensor slicing operator $:$ to denote simultaneous application across the temporal dimension, the spatial perturbation for all ROIs is formulated as:
\begin{equation}
X_s(r, :, c) = \alpha_c X(r, :, c) + \beta_c,
\end{equation}
where the scaling factor $\alpha_c \sim \mathcal{U}(\alpha_{min} , 1.0)$ and the offset $\beta_c \sim \mathcal{U}(-X_{\min}^{(c)}, P_{\max} - X_{\max}^{(c)})$, with $P_{\max}$ representing the maximum pixel value of the capture device.

\subsubsection{Gaussian Noise Injection}
To simulate sensor thermal noise globally across the entire spatial domain, bounded Gaussian perturbations are universally added to all ROIs. The augmented signal is formulated as:
\begin{equation}
X_g(r, :, c) = X(r, :, c) + \tilde{N}_{g}(r, :, c),
\end{equation}
where the additive noise $\tilde{N}_{g} \sim \mathcal{TN}(0, \sigma_g^2, -a_g, a_g)$ is drawn from a Truncated Normal distribution.

\subsubsection{Local Dropout}
To simulate transient occlusions, information is replaced within a random continuous temporal window $\mathcal{T}_r = [t_{start}, t_{start} + L_r - 1]$ for the selected regions. To ensure temporal validity, the window duration $L_r \sim \mathcal{U}\{1, T\}$ and the starting frame $t_{start} \sim \mathcal{U}\{1, T - L_r + 1\}$ are sampled from discrete uniform distributions. The perturbed signal is computed as:
\begin{equation}
X_d(r, t, c) =
\begin{cases}
\tilde{N}_{g}(t, c), & \text{if } r \in \mathcal{S}_d \text{ and } t \in \mathcal{T}_r \\
X(r, t, c), & \text{otherwise}
\end{cases}
\end{equation}
where $\mathcal{S}_d \subseteq \left[ \frac{R}{2} , R-1\right]$
denotes a subset of indices sampled uniformly at random.

Integrating the raw signal ($r$) alongside its perturbations, the expected loss is computed as:
\begin{equation}
\mathcal{L}_{smooth} = \sum_{i \in \{r, s, g, d\}} \lambda_i \mathcal{L}_{SNR, i} ,
\end{equation}
where $\lambda_i$ denotes the respective weighting coefficient.

\subsection{Noise-aware Null-space Regularization}
\label{NNR}
In real-world scenarios, rPPG signals are inevitably contaminated by complex noise due to their inherently low signal-to-noise ratio (SNR). To exploit noise interference as auxiliary optimization information, we propose a NNR module.

Specifically, based on the prior frequency range of BVP signals, denoted as $\Omega$, a zero-phase band-stop filter is applied along the temporal dimension to isolate the out-of-band noise representation $X_n$:
\begin{equation}
X_{n}(r, :, c) = \mathcal{H}_{bs}(X(r, :, c); \Omega),
\end{equation}
where $\mathcal{H}_{bs}(\cdot)$ denotes the zero-phase band-stop filter.

Subsequently, as illustrated in Fig. \ref{model} (a), we construct a noise-aware branch based on the LLP and WRD modules of the LSS.
To ensure the comprehensive retention of noise features, the extracted noise is projected into a space dimensionally equivalent to the physiological manifold.
This projection is formulated as:
\begin{equation}
Z_{n, i} = \mathcal{F}(X_{n, i}; \Theta_{nllp}),
\end{equation}
where $Z_{n,i} \in \mathbb{R}^{R \times C_l \times T}$ denotes the projected noise feature corresponding to the $i$-th spectral channel ($i \in \{R,G,B\}$), and $\Theta_{nllp}$ represents the projection parameters, sharing the LLP architecture but with independent weights. The features from all spectral channels are subsequently concatenated to obtain $Z_n \in \mathbb{R}^{R \times 3C_l \times T}$.

Concurrently, to guide the representation learning of $Z_n$, we introduce a WRD-based noise reconstruction task. The corresponding MSE loss is formulated as:
\begin{equation}
\mathcal{L}_{re} = \left\| X_n^{\prime} - X_n \right\|_F^2
\end{equation}
where $X_n^{\prime}=\mathcal{F}(Z_n; \Theta_{nwrd})$ is the reconstructed noise, and $\Theta_{nwrd}$ denotes the learnable parameters of the WRD-based reconstruction network.

To promote the disentanglement of the physiological representation $Z$ and the noise representation $Z_n$, an orthogonal loss $\mathcal{L}_{ortho}$ is introduced.
Minimizing the feature inner product explicitly constrains the physiological feature to evolve within the orthogonal null-space of the noise representation, thereby achieving effective low-dimensional feature purification.
The $\mathcal{L}_{ortho}$ is defined as:
\begin{equation}
\mathcal{L}_{ortho} = \frac{1}{R \cdot (3C_l)^2} \sum_{r=1}^{R} \sum_{i=1}^{3C_l} \sum_{j=1}^{3C_l} \left( \frac{\langle Z^{(r, i)}, Z_n^{(r, j)} \rangle}{\|Z^{(r, i)}\|_2 \|Z_n^{(r, j)}\|_2} \right)^2,
\end{equation}
where $Z^{(r, i)} \in \mathbb{R}^T$ denotes the vector of the $i$-th channel for the $r$-th ROI.

Crucially, the linear projection used in LLP prevents the feature entanglement commonly caused by complex non-linear distortions, ensuring a pristine decoupling of physiological and noise signals.

\subsection{Unsupervised Learning Paradigm}
\label{ulp}

In summary, integrating the aforementioned components yields the overall unsupervised training loss for the FCUS-rPPG framework, $\mathcal{L}_{us}$, formulated as:
\begin{equation}
\mathcal{L}_{us} = \sum_{i \in \{r, s, g, d\}} \lambda_i \left( \frac{1}{B - m + \epsilon} \sum_{b=1}^{B} M_b \mathcal{L}_{\text{SNR}, i}^{(b)} \right) + \lambda_{o} \mathcal{L}_{\text{ortho}}.
\label{eq:total_loss}
\end{equation}
Furthermore, the noise reconstruction loss $\mathcal{L}_{re}$ is jointly optimized to update the noise-aware branch, ensuring a robust noise representation.

\section{Experiments}
This section evaluates the proposed FCUS-rPPG framework on five datasets, including four public datasets, i.e., UBFC-rPPG \cite{bobbia2019unsupervised}, PURE \cite{stricker2014non}, BSIPL-motion \cite{Bian2025Motion}, and MMPD \cite{tang2023mmpd}, as well as one in-house dataset, BSIPL-RPPG \cite{song2021pulsegan}. Specifically, we conduct: 1) intra-dataset evaluation on UBFC-rPPG, PURE, and BSIPL-RPPG; 2) cross-dataset evaluation across all five datasets; and 3) comprehensive ablation studies of the proposed framework.

\subsection{Datasets}

\textbf{UBFC-rPPG} \cite{bobbia2019unsupervised} contains 42 video recordings acquired under real-world conditions.
Subjects were relatively stationary and performed a time-sensitive arithmetic task to elicit HR variability.
Videos were captured using a Logitech C920 HD Pro camera at 640×480 spatial resolution and 30 fps in uncompressed 8-bit RGB format.
Synchronous PPG signals were recorded through a Contec CMS50E pulse oximeter with 60 Hz sampling rate.

\textbf{PURE} \cite{stricker2014non} includes 60 video sequences from 10 subjects performing six distinct tasks: steady-state sitting, talking, slow horizontal head translation, fast horizontal head translation, small head rotations, and medium head rotations.
The videos were captured using an RGB ECO274CVGE camera (SVS-Vistek GmbH) at 640 × 480 pixel resolution and 30 fps.
GT PPG signals were concurrently recorded with a Contec CMS50E pulse oximeter attached to the subject’s finger.

\textbf{BSIPL-RPPG} \cite{song2021pulsegan} comprises 37 healthy student subjects (24 males, 13 females; ages 18–25 years).
Subjects sat 1 meter from a Logitech C920 HD Pro camera, while a Contec CMS50E pulse oximeter captured simultaneous PPG signals.
The video resolution was 640 × 480 pixels at 30 fps, and PPG data were recorded at a sampling rate of 60 Hz.
Each recording lasted approximately 4.5 minutes: the subjects were requested to sit still for the first 2 minutes, and perform some noticeable head movements for the last 2.5 minutes.

\textbf{MMPD} \cite{tang2023mmpd} includes 660 one-minute videos recorded using a Samsung Galaxy S22 Ultra mobile phone at 30 fps with a resolution of 1280 × 720 pixels, later downsampled to 320 × 240 pixels.
GT PPG signals were acquired using an HKG-07C+ oximeter at 200 Hz, then downsampled to 30 Hz.
The dataset encompasses a variety of conditions involving complex noises, including Fitzpatrick skin types 3–6, four lighting conditions (LED-low, LED-high, incandescent, natural), four activities (stationary, head rotation, talking, walking), and exercise scenarios.

\textbf{BSIPL-motion} \cite{Bian2025Motion} contains 264 one-minute video clips from 33 subjects performing four distinct motion scenarios: speaking, random rotation, horizontal movement, and anterior-posterior movement. The recordings were conducted under two lighting conditions (natural and LED light). Videos were captured using a Logitech C920 Pro HD webcam situated at 1.1 m from the subjects, at 640 × 480 spatial resolution and 30 fps in uncompressed YUV format. Synchronous PPG signals were recorded through a Contec CMS50E pulse oximeter with a 60 Hz sampling rate.

Table \ref{tab:dataset_summary} summarizes the scale of the five evaluated datasets by total video duration, which ranges from approximately 42 minutes (UBFC-rPPG) to 660 minutes (MMPD).

\begin{table}[t]
\centering
\caption{Statistical scale comparison of the datasets used in the experiments.}
\label{tab:dataset_summary}
\begin{tabular}{lccc}
\toprule
\textbf{Dataset} & \textbf{Subjects} & \textbf{Videos} & \textbf{Total (min)} \\
\midrule
UBFC-rPPG \cite{bobbia2019unsupervised}       & 42                   & 42              & 42             \\
PURE \cite{stricker2014non}             & 10                   & 60              & 60                   \\
BSIPL-RPPG \cite{song2021pulsegan}      & 37                   & 37              & 166.5          \\
BSIPL-motion \cite{Bian2025Motion}    & 33                   & 264             & 264                  \\
MMPD \cite{tang2023mmpd}            & 33                   & 660             & 660                  \\
\bottomrule
\end{tabular}
\end{table}

\begin{table*}[t]
\caption{Intra-dataset testing results on UBFC-rPPG, PURE, and BSIPL-RPPG.}
\label{intra_combined}
\centering
\scalebox{1}{
\setlength{\tabcolsep}{4pt}
\begin{tabular}{llccccccccc}
\toprule
\multirow{3}{*}{\centering Types} &
\multirow{3}{*}{\centering Methods} & \multicolumn{3}{c}{UBFC-rPPG} & \multicolumn{3}{c}{PURE} & \multicolumn{3}{c}{BSIPL-RPPG} \\
\cmidrule(lr){3-5} \cmidrule(lr){6-8} \cmidrule(lr){9-11}
& &
MAE (bpm) ↓ & RMSE (bpm) ↓ & $r$↑ &
MAE (bpm) ↓ & RMSE (bpm) ↓ & $r$↑ &
MAE (bpm) ↓ & RMSE (bpm) ↓ & $r$↑ \\
\midrule
\multirow{4}{*}{\rotatebox[origin=c]{90}{Tra.}}
& GREEN \cite{verkruysse2008remote}   & 10.09 & 23.85 & 0.34 & 19.73 & 31.00 & 0.37 & 22.48 & 28.38 & -0.17 \\
& ICA \cite{poh2010non}               & 4.77  & 16.07 & 0.72 & 16.00 & 25.65 & 0.44 & 13.31 & 20.86 & 0.07 \\
& CHROM \cite{de2013robust}           & 5.77  & 14.93 & 0.81 & 4.06  & 8.83  & 0.89 & 3.28  & 6.93  & 0.85 \\
& POS \cite{wang2016algorithmic}      & 3.67  & 11.82 & 0.88 & 4.08  & 7.72  & 0.92 & \underline{2.34} & \underline{4.62} & \underline{0.93} \\
\midrule
\multirow{6}{*}{\rotatebox[origin=c]{90}{Sup.}}
& HR-CNN \cite{vspetlik2018visual}    & -     & -     & -    & 1.84  & 2.37  & 0.98 & - & - & - \\
& RPNet \cite{speth2021unifying}      & 0.53  & 1.78  & \textbf{0.99} & 1.15 & 5.77 & 0.96 & - & - & - \\
& DeepPhys \cite{chen2018deepphys} & - & - & - & - & - & - & 25.72 & 28.30 & 0.10\\
& EfficientPhys \cite{Liu_2023_WACV} & - & - & - & -  & - &- & 24.21 & 27.30 & 0.18\\
& PhysNet \cite{yu2019remote}         & 0.55  & 2.03  & \textbf{0.99} & 0.99 & 5.22 & 0.97 & 7.74 & 12.45 & 0.42 \\
& Dual-GAN \cite{lu2021dual}          & 0.44  & \underline{0.67} & \textbf{0.99} & 0.82 & 1.31 & \underline{0.99} & - & - & - \\
\midrule
\multirow{9}{*}{\rotatebox[origin=c]{90}{L-F.}}
& Gideon2021 \cite{gideon2021way}      & 1.85  & 4.28  & 0.93 & 2.30 & 2.90 & \underline{0.99} & - & - & - \\
& SLF-RPM \cite{wang2022slfrpm}        & 8.39  & 9.70  & 0.70 & -    & -    & -    & - & - & - \\
& SimPer \cite{yang2022simper}         & 4.24  & -     & -    & 3.89 & -    & -    & - & - & - \\
& Contrast-Phys+ \cite{sun2024contrastplus} & 0.64 & 1.00 & \textbf{0.99} & 1.00 & 1.40 & \underline{0.99} & 5.84 & 17.00 & 0.65 \\
& rPPG-MAE \cite{liu2024rppgmae}       & \textbf{0.17} & \textbf{0.21} & \textbf{0.99} & \textbf{0.40} & \textbf{0.92} & \underline{0.99} & - & - & - \\
& SiNC \cite{speth2023non}             & 0.59  & 1.83  & \textbf{0.99} & 0.61 & 1.84 & \textbf{1.00} & - & - & - \\
& Yue \textit{et al}. \cite{yue2023facial} & 0.58 & 0.94 & \textbf{0.99} & 1.23 & 2.01 & \underline{0.99} & - & - & - \\
& Ma \textit{et al}. \cite{MA2026113501} & 0.38 & 0.92 & \textbf{0.99} & 0.60 & 1.18 & \underline{0.99} & - & - & - \\
& \textbf{FCUS-rPPG (Ours)}                        & \underline{0.31} & 1.03 & \textbf{0.99} & \underline{0.49} & \underline{1.10} & \underline{0.99} & \textbf{1.32} & \textbf{3.20} & \textbf{0.96} \\
\bottomrule
\end{tabular}
}

\vspace{1mm}
\parbox{0.95\linewidth}{\footnotesize Best results are \textbf{bold}, second best are \underline{underlined}. "Tra.", "Sup.", and "L-F." denote traditional, supervised, and label-free methods, respectively.}
\end{table*}

\begin{figure*}[t]
\centering
\includegraphics[width=1\linewidth]{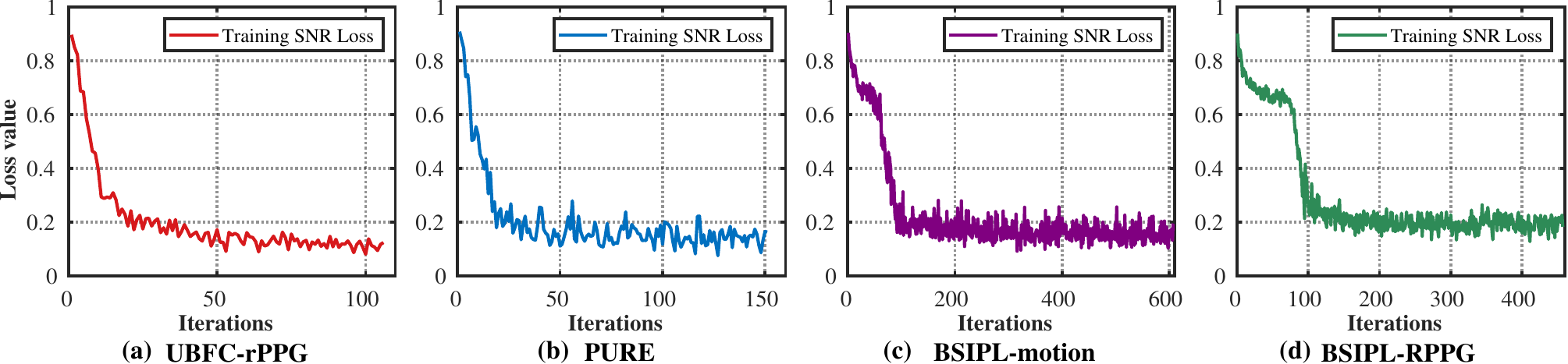}
\caption{SNR loss curves of raw signals during single-epoch training across different datasets: (a) UBFC-rPPG, (b) PURE, (c) BSIPL-motion, and (d) BSIPL-RPPG. The number of iterations within one training epoch varies across datasets, with 106 iterations for UBFC-rPPG, 150 iterations for PURE, 607 iterations for BSIPL-motion, and 456 iterations for BSIPL-RPPG.}
\label{oneepochloss}
\end{figure*}

\begin{table*}[t]
\caption{Cross-dataset testing results on three datasets: PURE, UBFC-rPPG, and BSIPL-motion.}
\label{crossresult1}
\centering
\scalebox{1}{
\setlength{\tabcolsep}{3pt} 
\begin{tabular}{llccccccccc}
\toprule
\multirow{3}{*}{\centering Types} &
\multirow{3}{*}{\centering Methods} & \multicolumn{3}{c}{UBFC-rPPG$\rightarrow$PURE} & \multicolumn{3}{c}{PURE$\rightarrow$UBFC-rPPG} & \multicolumn{3}{c}{UBFC-rPPG$\rightarrow$BSIPL-motion} \\
\cmidrule(lr){3-5} \cmidrule(lr){6-8} \cmidrule(lr){9-11}
& &
MAE (bpm) ↓ & RMSE (bpm) ↓ & $r$↑ &
MAE (bpm) ↓ & RMSE (bpm) ↓ & $r$↑ &
MAE (bpm) ↓ & RMSE (bpm) ↓ & $r$↑ \\
\midrule
\multirow{9}{*}{\rotatebox[origin=c]{90}{Sup.}}
& DeepPhys \cite{chen2018deepphys} & 5.54 & 18.51 & 0.66 & 1.21 & 2.90 & \textbf{0.99} & 14.63 & 19.68 & 0.12\\
&RhythmNet \cite{niu2019rhythmnet}  & 7.39 & 10.49 & 0.77 & 5.79 & 7.91 & 0.78 & - & - & -\\
& TS-CAN \cite{liu2020multi} & 3.69 & 13.80 & 0.82 & 1.30 & 2.87 & \textbf{0.99} & 10.87 & 16.07 & 0.34\\
& EfficientPhys \cite{Liu_2023_WACV} & 5.47 & 17.04 & 0.71 & 2.07  & 6.32 & 0.94 & 11.59 & 17.34 & 0.24\\
& PhysNet \cite{yu2019remote} & 8.06 & 19.71 & 0.61 & \underline{0.98} & \underline{2.48} & \textbf{0.99} & \underline{7.50} & \underline{12.91} & \underline{0.41}\\
& Physformer \cite{yu2022physformer} & 12.92 & 24.36 & 0.47 & 3.96 & 13.57 & 0.89 & 11.04 & 16.11 & 0.21 \\
& PulseGAN \cite{song2021pulsegan} & 3.36 & 5.11 & 0.95 & 2.30 & 3.50 & \underline{0.97} & - & - & -\\
& NEST \cite{Lu2023NEST} & 6.07 & 9.06 & 0.76 & 4.67 & 6.79 & 0.86 & - & - & -\\
& Greip \cite{zhang2025agrtip} & 5.70 & 8.23 & 0.88 & 4.08 & 6.17 & 0.88 & - & - & -\\
\midrule
\multirow{7}{*}{\rotatebox[origin=c]{90}{L-F.}}
& Gideon2021 \cite{gideon2021way} & 2.95 & 4.60 & 0.97 & 2.37 & 3.51 & 0.95 & - & - & -\\
& rPPG-MAE \cite{liu2024rppgmae} & 13.55 & 20.27 & - & 1.28 & 2.75 & - & - & - & - \\
& Contrast-Phys+ \cite{sun2024contrastplus} & 2.84 & 11.87 & 0.87 & - & - & - & 13.60 & 19.11 & 0.34\\
& SiNC \cite{speth2023non} & 4.02 & - & 0.86 & 6.64 & - & 0.59 & 10.16 & 17.08 & 0.30\\
& Yue \textit{et al}. \cite{yue2023facial} & \underline{2.14} & \underline{3.37} & \underline{0.98} & 2.18 & 3.20 & 0.97 & - & - & -\\
& Ma \textit{et al}. \cite{MA2026113501} & 11.45 & 24.78 & - & 1.51 & 3.17 & - & - & - & -\\
& \textbf{FCUS-rPPG (Ours)}  & \textbf{0.49} & \textbf{1.24} & \textbf{0.99} & \textbf{0.71} & \textbf{2.16} & \textbf{0.99} & \textbf{0.55} & \textbf{1.33} & \textbf{0.99}  \\
\bottomrule
\end{tabular}
}

\vspace{1mm}
\parbox{0.95\linewidth}{\footnotesize Best results are \textbf{bold}, second best are \underline{underlined}.}
\end{table*}

\begin{table*}[t]
\caption{Cross-dataset testing results on three datasets: MMPD, BSIPL-rPPG, and PURE.}
\label{crossresult2}
\centering
\scalebox{1}{
\setlength{\tabcolsep}{3pt} 
\begin{tabular}{llccccccccc}
\toprule
\multirow{3}{*}{\centering Types} &
\multirow{3}{*}{\centering Methods} & \multicolumn{3}{c}{UBFC-rPPG$\rightarrow$MMPD} & \multicolumn{3}{c}{UBFC-rPPG$\rightarrow$BSIPL-RPPG} & \multicolumn{3}{c}{BSIPL-motion$\rightarrow$PURE} \\
\cmidrule(lr){3-5} \cmidrule(lr){6-8} \cmidrule(lr){9-11}
& &
MAE (bpm) ↓ & RMSE (bpm) ↓ & $r$↑ &
MAE (bpm) ↓ & RMSE (bpm) ↓ & $r$↑ &
MAE (bpm) ↓ & RMSE (bpm) ↓ & $r$↑ \\
\midrule
\multirow{5}{*}{\rotatebox[origin=c]{90}{Sup.}}
& DeepPhys \cite{chen2018deepphys} & 15.02 & 22.69 & 0.18 & 9.47 & 17.13 & 0.19 & 20.85 & 30.97 & 0.04\\
& EfficientPhys \cite{Liu_2023_WACV} & 14.19 & 22.21 & 0.22 & - & - & - & 21.10 & 30.91 & 0.00\\
& TS-CAN \cite{liu2020multi} & 14.70 & 21.97 & 0.20 & 7.14 & 15.23 & 0.22 & - & - & -\\
& PhysNet \cite{yu2019remote} & \underline{10.97} & \underline{17.35} & \underline{0.31} & 5.26 & 12.48 & 0.48 & 9.33 & 20.14 & \underline{0.59}\\
& Physformer \cite{yu2022physformer} & 12.64 & 18.41 & 0.20 & 6.38 & 13.77 & 0.39 & 18.71 & 25.67 & 0.00 \\
\midrule
\multirow{3}{*}{\rotatebox[origin=c]{90}{L-F.}}
& SiNC \cite{speth2023non} & - & - & - & 3.03 & 7.86 & 0.83  & \underline{7.41} & \underline{18.58} & 0.58\\
& Contrast-Phys+ \cite{sun2024contrastplus} & 13.33 & 23.52 & 0.25 & \underline{2.15} & \underline{6.41} & \underline{0.88} & 13.63 & 24.45 & 0.33\\
& \textbf{FCUS-rPPG (Ours)}   & \textbf{8.96} & \textbf{15.18} & \textbf{0.42} & \textbf{1.61} & \textbf{4.27} & \textbf{0.95} & \textbf{0.60} & \textbf{1.43} & \textbf{0.99}  \\
\bottomrule
\end{tabular}
}

\vspace{1mm}
\parbox{0.95\linewidth}{\footnotesize Best results are \textbf{bold}, second best are \underline{underlined}.}
\end{table*}

\subsection{Experimental Setup and Metrics}
During the training stage, each video is segmented into 15-second clips with a 1-second stride as input.
In the testing phase, we follow the protocol in \cite{sun2024contrastplus}, dividing each video into non-overlapping 30-second segments and computing the average HR from the predicted BVP signal per segment.

The physiological frequency band $\Omega$ is set to $[0.66, 3]$ Hz following \cite{speth2023non}, and the frequency tolerance $\Delta f$ is set to 0.3 Hz. The amplitude threshold $\tau$ is fixed at 1 and can be adjusted according to the acquisition device \cite{verkruysse2008remote}. The numerical stability constant $\epsilon$ is set to $10^{-8}$.
For data perturbation, we adopt highly randomized hyperparameters to improve optimization robustness. Specifically, $\alpha_{min}=0.5$, $a_g \sim \mathcal{U}(0,1)$, and $\sigma_g \sim \mathcal{U}(1,2)$. All weighting coefficients $\lambda_i(i \in \{r, s, g, d\})$ are set to 1, while $\lambda_o$ is set to 0.1.

Our proposed method is implemented in PyTorch and trained on two NVIDIA TITAN Xp GPUs.
The model parameters are initialized with Kaiming normal initialization \cite{he2015delving}.
We optimize the model using the AdamW optimizer with a learning rate of 0.01 and a batch size of 20.
Across all evaluated datasets, the model is trained for a single epoch.

Following previous studies \cite{speth2023non, yue2023facial, liu2024rppgmae, sun2024contrastplus}, we evaluate the average HR using the most common performance metrics, including mean absolute error (MAE), root mean square error (RMSE), and Pearson correlation coefficient (r).

\subsection{Intra-Dataset Testing}
For intra-dataset evaluation, we select three relatively small-scale datasets: UBFC-rPPG, PURE, and BSIPL-RPPG. To ensure fair comparisons, we strictly follow the widely adopted protocols in previous studies. Specifically, the standard split (30 training and 12 testing videos) is used for UBFC-rPPG \cite{sun2022contrast, song2021pulsegan}, while 5-fold subject-independent cross-validation is employed for PURE and BSIPL-RPPG \cite{niu2020video, yue2023facial}.

Table \ref{intra_combined} presents the quantitative HR estimation results. Three observations can be drawn from these experiments. First, compared with traditional methods, DL-based methods generally achieve superior performance, demonstrating stronger capability in recovering BVP signals when the noise distribution aligns with the training data. Second, among DL-based methods, unsupervised frameworks often exhibit stronger robustness than supervised methods, confirming the efficacy of unsupervised learning as a promising alternative to label-dependent baselines. Third, the proposed FCUS-rPPG achieves highly competitive performance compared with existing SOTA unsupervised approaches. Specifically, our method yields an MAE of 0.31 bpm on UBFC-rPPG and 0.49 bpm on PURE, closely approaching the SOTA rPPG-MAE (0.17 bpm and 0.40 bpm, respectively).

\subsection{Cross-Dataset Testing}

\begin{figure}[t]
\centering
\includegraphics[width=1 \linewidth]{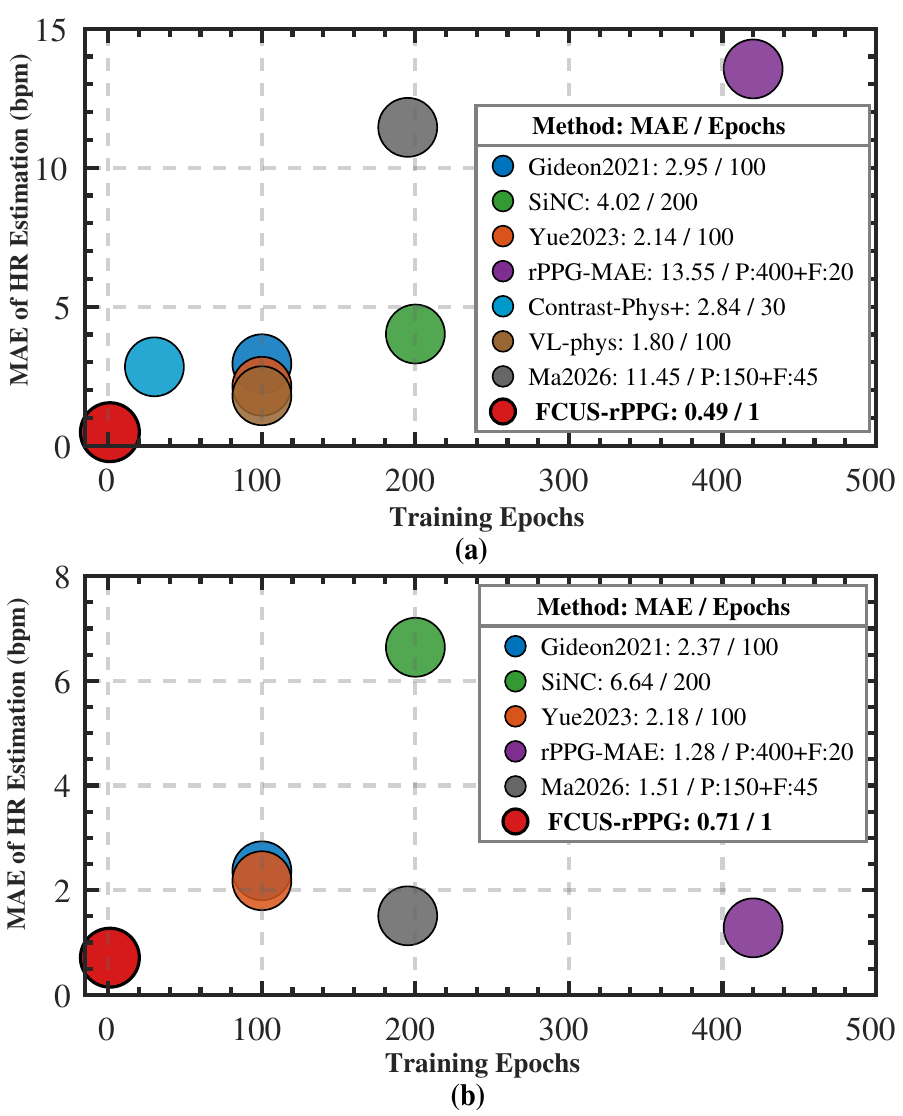}
\caption{Cross-dataset evaluation of the accuracy-epochs tradeoff across different methods: (a) trained on UBFC-rPPG and tested on PURE, and (b) trained on PURE and tested on UBFC-rPPG. The y-axis represents the MAE of HR estimation, and the x-axis denotes the training epochs ('P' for pre-training, 'F' for fine-tuning). The method in the bottom-left corner exhibits the optimal accuracy-efficiency tradeoff.}
\label{tradeoff}
\end{figure}

\begin{figure}[t]
\centering
\includegraphics[width=1 \linewidth]{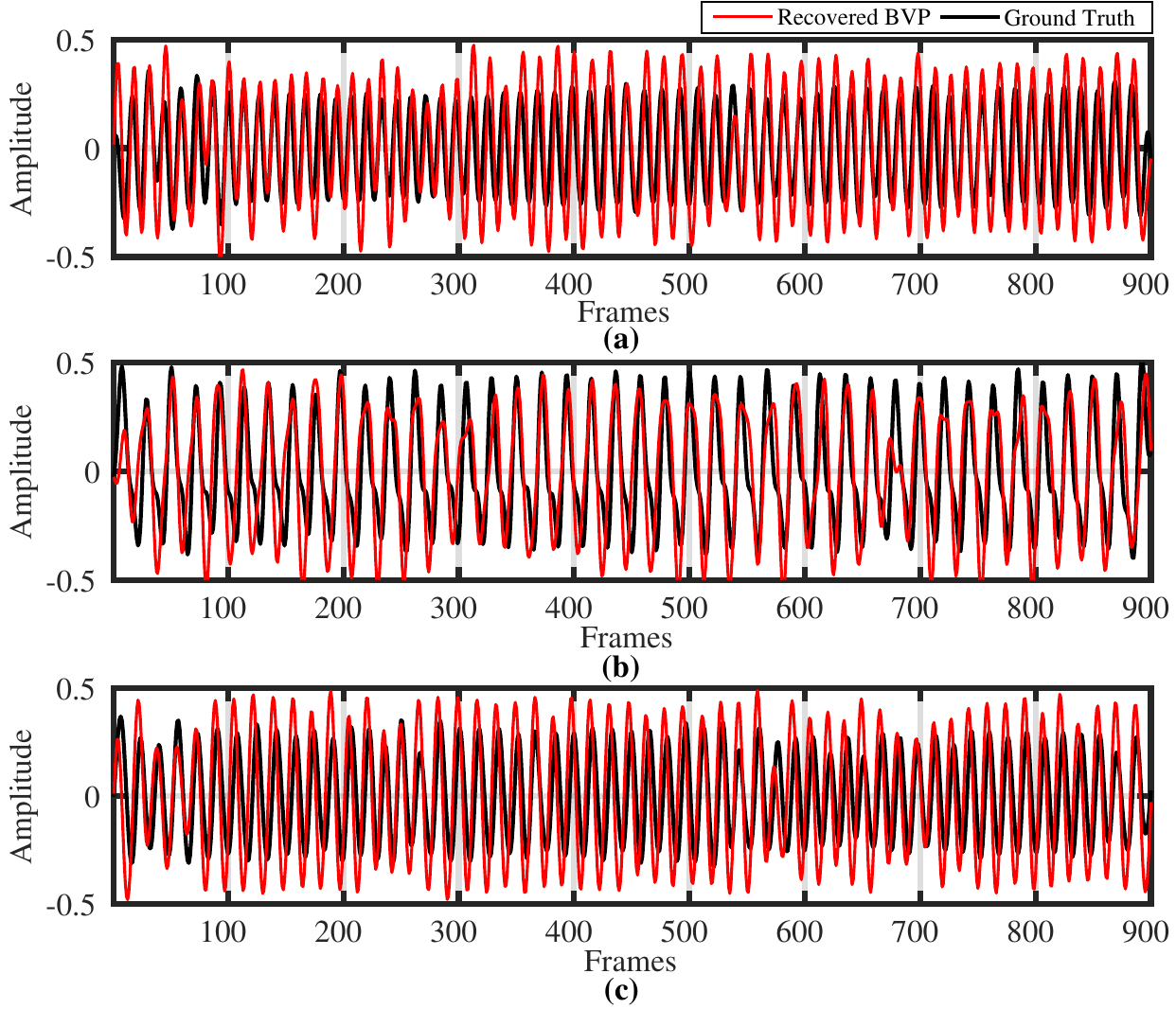}
\caption{Visualizations of BVP signals recovered by the FCUS-rPPG framework under cross-dataset testing scenarios: (a) trained on UBFC-rPPG and tested on PURE; (b) trained on UBFC-rPPG and tested on BSIPL-motion; and (c) trained on PURE and tested on UBFC-rPPG.}
\label{wave}
\end{figure}

To evaluate the generalization capability of the proposed framework, we conduct comprehensive cross-dataset evaluations on five datasets: UBFC-rPPG, PURE, BSIPL-RPPG, BSIPL-motion, and MMPD.
Our primary protocol uses UBFC-rPPG, which contains relatively simple noise distributions, for training, while evaluations are conducted on more challenging datasets featuring severe motion artifacts and illumination variations.
Furthermore, to verify convergence robustness under challenging optimization conditions, we also train the model on datasets with complex noise profiles (PURE and BSIPL-motion).
The results of Contrast-Phys+ (0\%) \cite{sun2024contrastplus} and SiNC \cite{speth2023non} are reproduced using their official codes.
The remaining baseline results are either cited from their original publications \cite{yue2023facial, speth2023non, MA2026113501} or reproduced using the open-source rPPG-Toolbox \cite{liu2023rppgtool}.

As illustrated by the loss curves in Fig. \ref{oneepochloss}, the proposed method demonstrates rapid and stable convergence within a single epoch.
Notably, this convergence behavior remains stable on both limited-scale datasets (UBFC-rPPG) and those corrupted by severe noise and motion artifacts (PURE, BSIPL-motion, BSIPL-rPPG).

Tables \ref{crossresult1} and \ref{crossresult2} detail the quantitative results for cross-dataset HR estimation.
Based on these evaluations, we draw three primary conclusions.
First, most existing DL baselines exhibit substantial performance degradation under cross-dataset evaluation compared with their intra-dataset results.
This observation suggests that existing optimization schemes tend to converge toward dataset-specific sharp minima that generalize poorly under distribution shifts.
Second, the proposed FCUS-rPPG achieves consistently SOTA performance across all evaluated datasets.
Unlike existing DL-based methods, our framework shows minimal performance decay during cross-dataset testing, demonstrating highly robust, domain-invariant BVP feature extraction.
Third, as illustrated in Table \ref{crossresult2}, several DL baselines fail to converge when trained on datasets corrupted by severe noise (e.g., BSIPL-motion), resulting in $r=0$ on the test set.
In contrast, FCUS-rPPG maintains stable optimization trajectories and achieves reliable convergence behavior even under severe noise interference.

Fig. \ref{tradeoff} further illustrates the trade-off between estimation error (MAE) and computational cost (training epochs) for the bidirectional cross-dataset evaluations between PURE and UBFC-rPPG.
FCUS-rPPG consistently achieves the minimum estimation error with the fewest training epochs.
This demonstrates a superior balance between training efficiency and generalization capacity.

Fig. \ref{wave}  visualizes the recovered BVP waveforms under cross-dataset testing scenarios. Notably, despite the absence of GT labels during training, the waveforms reconstructed by FCUS-rPPG exhibit a high degree of consistency with the GT PPG signals, even on datasets corrupted by complex noise.
These results confirm that the proposed framework not only yields highly generalizable average HR estimations but also demonstrates a robust capability for high-quality BVP signal reconstruction.

\subsection{Ablation Studies}
\subsubsection{Impact of Low-dimensional Spectrally-shared Backbone}
\begin{figure}[t]
\centering
\includegraphics[width=1\linewidth]{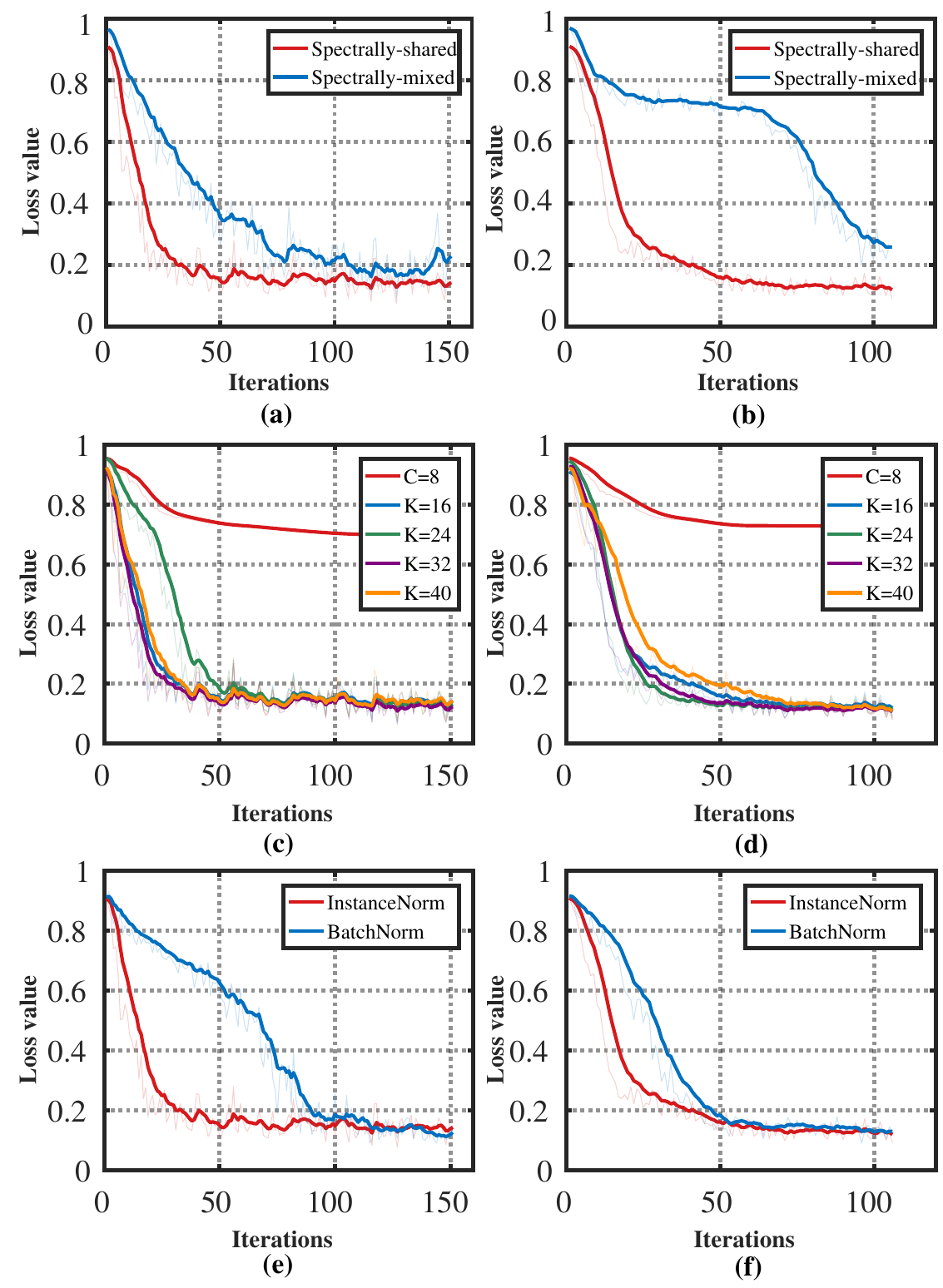}
\caption{Convergence curves of the raw input SNR loss during training on the PURE ((a), (c), (e)) and UBFC-rPPG ((b), (d), (f)) datasets. The subfigures present ablation results for three key components: (a) and (b) spectrally-shared versus spectrally-mixed modeling; (c) and (d) feature channel configurations; and (e) and (f) normalization strategies.}
\label{ab_LSS}
\end{figure}

To validate the effectiveness of the proposed low-dimensional spectrally-shared backbone (LSS), we conduct ablation studies on the UBFC-rPPG and PURE datasets from two perspectives: convergence behavior (Fig. \ref{ab_LSS}) and cross-dataset generalization performance (Table \ref{tab:ab_LSS}).

First, to evaluate the impact of spectrally-shared modeling, we compare two approaches: spectrally-shared and spectrally-mixed (i.e., mapping the joint 3-channel input to a final $3C_l$-dimensional feature space).
As shown in Fig. \ref{ab_LSS} (a) and (b), the parameter-sharing strategy converges significantly faster than the mixed approach. Table \ref{tab:ab_LSS} further confirms its superior estimation accuracy.
These results corroborate our hypothesis (Section \ref{mpc}) regarding the spectral physiological covariation inherent in rPPG signals.

Second, we investigate the impact of the feature channel dimension, $C_l$, by increasing it from 8 to 40 in steps of 8.
As shown in Fig. \ref{ab_LSS} (c) and (d), the model strictly diverges on both datasets when $C_l=8$.
Conversely, configuring $C_l \geq 16$ ensures extremely rapid convergence.
Table \ref{tab:ab_LSS} reveals that expanding the channel capacity beyond 16 yields no further performance gains and even causes slight degradation.
This substantiates our premise (Section \ref{lpmh}) that BVP features reside within a low-dimensional manifold, rendering $C_l=16$ sufficient for comprehensive representation.

Finally, to determine the optimal normalization strategy, we ablate the normalization operations within the network, focusing on the efficacy of BN versus IN.
The experimental results presented in Fig. \ref{ab_LSS} (e), (f), and Table \ref{tab:ab_LSS} clearly indicate that, compared to BN, which heavily relies on the training data scale, adopting the IN strategy not only substantially accelerates the convergence rate but also secures a significant advantage in final estimation accuracy.

\begin{table}[t]
\caption{Ablation experiments on low-dimensional spectrally-shared backbone.}
\label{tab:ab_LSS}
\centering
\setlength{\tabcolsep}{1.5pt}
\begin{tabularx}{\columnwidth}{lYYYYYY}
\toprule
\multirow{2}{*}{\textbf{Configuration}} &
\multicolumn{3}{c}{PURE $\rightarrow$ UBFC-rPPG} &
\multicolumn{3}{c}{UBFC-rPPG $\rightarrow$ PURE}
\\
\cmidrule(lr){2-4} \cmidrule(lr){5-7}
& MAE$\downarrow$ (bpm) & RMSE$\downarrow$ (bpm) & $r\uparrow$ & MAE$\downarrow$ (bpm) & RMSE$\downarrow$ (bpm) & $r\uparrow$
\\ \midrule

\rowcolor[gray]{0.95} \multicolumn{7}{l}{\textit{Impact of Spectral Modeling Strategies}} \\
Spectrally-mixed    & 1.30 & 3.55 & 0.97 & 0.54 & 1.36 & 0.99 \\
spectrally-shared  & \textbf{0.71} & \textbf{2.16} & \textbf{0.99} & \textbf{0.49} & \textbf{1.24} & \textbf{0.99} \\

\midrule
\rowcolor[gray]{0.95} \multicolumn{7}{l}{\textit{Impact of Channel Numbers (C)}} \\
C=8  & 49.05 & 52.01 & 0.00 & 20.39 & 30.49 & 0.00 \\
C=16 & \textbf{0.71} & \textbf{2.16} & \textbf{0.99} & \textbf{0.49} & \textbf{1.24} & \textbf{0.99} \\
C=24 & 0.71 & 2.16 & 0.99 & 0.52 & 1.33 & 0.99 \\
C=32 & 0.84  & 2.75  & 0.99 & 0.49  & 1.29 & 0.99 \\
C=40 & 0.87  & 2.66  & 0.99 & 0.60  & 1.45 & 0.99 \\

\midrule
\rowcolor[gray]{0.95} \multicolumn{7}{l}{\textit{Impact of Normalization Layers}} \\
BatchNorm    & 2.14 & 5.95 & 0.94 & 1.17 & 6.11 & 0.96 \\
InstanceNorm  & \textbf{0.71} & \textbf{2.16} & \textbf{0.99} & \textbf{0.49} & \textbf{1.24} & \textbf{0.99} \\

\bottomrule
\end{tabularx}

\vspace{1mm}
\parbox{\linewidth}{\footnotesize 
$\downarrow$ indicates lower is better, $\uparrow$ indicates higher is better. Best results are \textbf{bold}.
}
\end{table}

\subsubsection{Impact of Post-verification Gradient Masking}
\begin{figure}[t]
\centering
\includegraphics[width=1\linewidth]{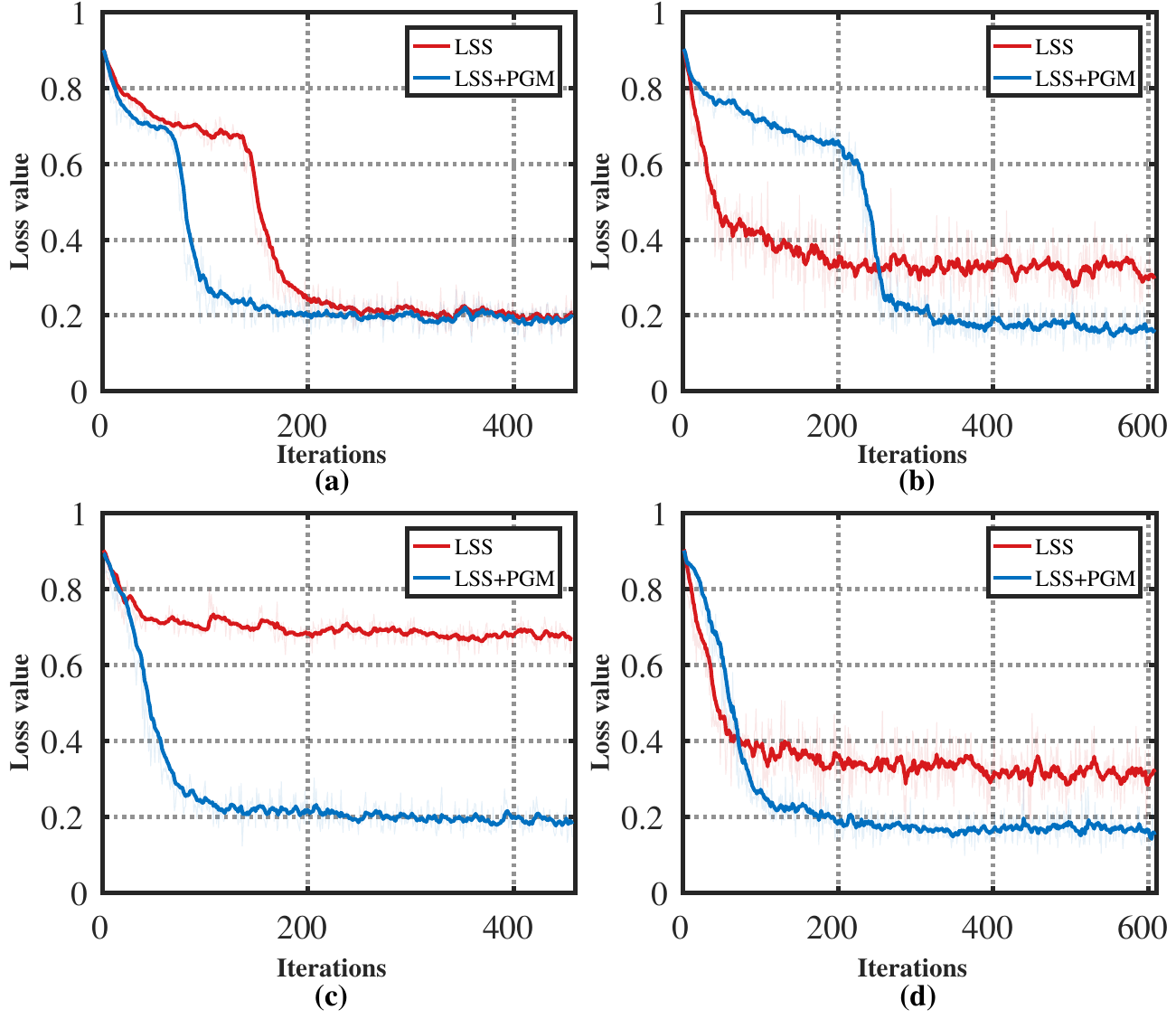}
\caption{Convergence curves of the SNR loss under two different random data loading seeds on the BSIPL-RPPG ((a), (c)) and BSIPL-motion ((b), (d)) datasets.}
\label{ab_PGM}
\end{figure}
To evaluate the contribution of the PGM module in mitigating gradient contamination, the convergence behavior of the base framework (LSS) is compared against the PGM-enhanced framework (LSS+PGM).
Fig. \ref{ab_PGM} illustrates the training loss curves on two complex noisy datasets (BSIPL-motion and BSIPL-rPPG) across two random data loading seeds.
The results demonstrate two primary advantages of the PGM mechanism.
First, it rescues divergent optimization trajectories. Under the setting of data seed 1 (Fig. \ref{ab_PGM} (a) and (b)), the base framework manages to converge on BSIPL-rPPG but completely diverges on BSIPL-motion, which contains severe motion artifacts. When switching to seed 2 (Fig. \ref{ab_PGM} (c) and (d)), the base framework encounters optimization collapse on both datasets. In contrast, integrating the PGM module enables the framework to achieve stable convergence under both random loading seeds. This demonstrates that the PGM effectively filters out gradient contamination induced by periodic noise.
Second, it significantly enhances convergence efficiency. As shown in Fig. \ref{ab_PGM} (a), even in scenarios where the base framework originally manages to converge, the introduction of the PGM drastically curtails the number of epochs required to reach convergence, yielding a marked leap in optimization efficiency.
\subsubsection{Impact of Loss Landscape Smoothing}
\begin{figure}[t]
\centering
\includegraphics[width=1\linewidth]{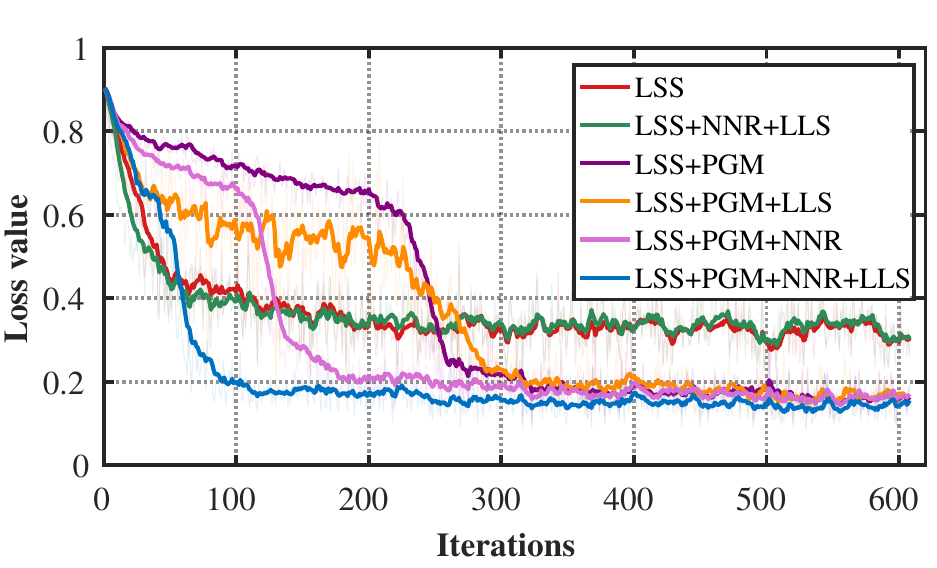}
\caption{Convergence curves of the raw input SNR loss during training on the BSIPL-motion.}
\label{AB}
\end{figure}

\begin{table}[t]
\caption{Ablation experiments on loss landscape smoothing.}
\label{tab:ab_LLS}
\centering
\setlength{\tabcolsep}{1.5pt}
\begin{tabularx}{\columnwidth}{lYYYYYY}
\toprule
\multirow{2}{*}{\textbf{Configuration}} &
\multicolumn{3}{c}{BSIPL-motion $\rightarrow$ PURE} &
\multicolumn{3}{c}{BSIPL-RPPG $\rightarrow$ BSIPL-motion}
\\
\cmidrule(lr){2-4} \cmidrule(lr){5-7}
& MAE$\downarrow$ (bpm) & RMSE$\downarrow$ (bpm) & $r\uparrow$ & MAE$\downarrow$ (bpm) & RMSE$\downarrow$ (bpm) & $r\uparrow$ \\

\midrule
w/o LLS & 0.67 & 1.61 & 0.99 & 0.73 & 2.93 & 0.95 \\
w/ LLS  & \textbf{0.57} & \textbf{1.46} & \textbf{0.99} & \textbf{0.64} & \textbf{2.43} & \textbf{0.97} \\

\bottomrule
\end{tabularx}

\vspace{1mm}
\parbox{\linewidth}{\footnotesize
$\downarrow$ indicates lower is better, $\uparrow$ indicates higher is better. Best results are \textbf{bold}.
}
\end{table}

To evaluate the effectiveness of the LLS mechanism, comparative ablation experiments are conducted by isolating the LLS module within the otherwise complete framework.
Table \ref{tab:ab_LLS} and Fig. \ref{AB} illustrate the specific impacts of LLS on generalization performance and convergence speed, respectively.
As shown in Table \ref{tab:ab_LLS}, cross-dataset evaluations on the BSIPL-motion and PURE datasets indicate that incorporating the LLS significantly improves the generalization capabilities of the converged model.
Furthermore, the loss curves in Fig. \ref{AB} demonstrate that the complete framework (LSS+PGM+NNR+LLS) converges considerably faster than its ablated counterpart (LSS+PGM+NNR).
These results confirm that the LLS not only guides the model toward a flat minima space for superior generalization but also inherently stabilizes gradient updates and accelerates optimization.
\subsubsection{Impact of Noise-aware Null-space Regularization}
To verify the contribution of the NNR mechanism to model optimization, systematic ablation experiments are conducted on the BSIPL-motion dataset, which features complex motion noise (Fig. \ref{AB}).
Given that the PGM module ensures basic convergence, the network's convergence dynamics are evaluated under two comparative settings: (1) the complete framework (LSS+PGM+NNR+LLS) versus its NNR-ablated counterpart (LSS+PGM+LLS), and (2) the NNR-equipped baseline (LSS+PGM+NNR) versus the basic configuration (LSS+PGM).
Both comparisons consistently demonstrate that integrating the NNR module significantly accelerates convergence compared to the respective baselines.
These findings confirm that the NNR can impose effective spatial constraints on the optimization trajectory, thereby substantially enhancing convergence efficiency under complex noise interference.

\subsection{Discussion}

\begin{table}[t]
\caption{Comparison of computational cost and performance across different rPPG methods trained on the UBFC-rPPG dataset and tested on the PURE dataset.}
\label{tab:pam_performance}
\centering
\setlength{\tabcolsep}{4pt}
\begin{tabularx}{\linewidth}{@{} X c c c c c @{}}
\toprule
\multirow{2}{*}{Method} & Pre-process & Train & \multirow{2}{*}{Params} & \multicolumn{2}{c}{Error (bpm) $\downarrow$} \\
\cmidrule(lr){5-6}
 & (m:s) & (h:m:s) & & MAE & RMSE \\
\midrule
DeepPhys$^*$ \cite{wang2016algorithmic}             & 13:08 & 00:12:30 & 2.23M & 5.54 & 18.51 \\
EfficientPhys$^*$ \cite{Liu_2023_WACV}              & 11:51 & 00:11:46 & 2.16M & 5.47 & 17.04 \\
PhysNet$^*$ \cite{yu2019remote}                     & 10:52 & 00:12:06 & 0.77M & 8.06 & 19.71 \\
\midrule
Contrast-Phys+$^\dagger$ \cite{sun2024contrastplus} & 08:53 & 01:10:02 & 0.86M & 2.84 & 11.87 \\
SiNC$^\dagger$ \cite{speth2023non}                  & 30:12 & 04:05:04 & 1.38M & 4.02 & --    \\
\midrule
FCUS-rPPG (Ours)$^\dagger$                          & \textbf{05:50} & \textbf{00:00:40} & \textbf{8.5K} & \textbf{0.49} & \textbf{1.24} \\
\bottomrule
\end{tabularx}

\vspace{1mm}
\parbox{\linewidth}{\footnotesize Best results are highlighted in \textbf{bold}. "h", "m", and "s" denote hours, minutes, and seconds, respectively. $^*$ indicates supervised methods, and $^\dagger$ indicates label-free methods. To ensure a fair comparison, the reported computational times are evaluated on the same device.
}
\end{table}

The efficacy of the FCUS-rPPG stems from precise theoretical modeling and synergistic optimization strategies.
Although BVP amplitudes vary significantly across subjects, devices, and illumination conditions, camera-based physiological signals share a common physical property: they contain subtle quasi-periodic physiological variations embedded within noise.
This property constrains physiological representations to a low-dimensional manifold, motivating the proposed low-dimensional spectrally-shared parameterization strategy.
From an optimization perspective, the key challenge in unsupervised rPPG lies in transforming physiological priors into effective optimization constraints. The proposed framework explicitly incorporates physiological subtlety, perturbation diversity, and noise-feature orthogonality into the optimization process.

As shown in Table \ref{tab:pam_performance}, FCUS-rPPG achieves SOTA cross-dataset generalization while requiring substantially fewer model parameters and remarkably shorter training time.
Achieving ultra-fast from-scratch convergence has important implications for practical rPPG deployment.
This work provides both theoretical insights and experimental validation for the feasibility of rapid unsupervised training with strong generalization capability.
In essence, the proposed paradigm offers a viable approach for rapid, direct adaptation from unlabeled facial videos without requiring any pre-training.
\section{Conclusion}
In this work, we propose FCUS-rPPG, the first unsupervised rPPG framework that simultaneously achieves rapid optimization and robust generalization.
By revisiting rPPG representation learning from the perspective of multi-spectral physiological covariation and low-dimensional manifold structure, we develop a low-dimensional spectrally-shared backbone that effectively reduces optimization complexity while learning domain-invariant physiological representations. 
To further improve convergence efficiency and gradient reliability, we develop a unified optimization framework consisting of amplitude-prior-guided post-verification gradient masking, perturbation-based loss landscape smoothing, and noise-aware null-space regularization.
Extensive experiments on five datasets demonstrate that FCUS-rPPG achieves superior cross-dataset generalization with only a single training epoch from scratch under identical dataset settings.
In future work, we will explore ultra-fast adaptation methods for personalized physiological monitoring, relying solely on limited unlabeled facial videos to facilitate rapid on-device deployment.


\bibliographystyle{IEEEtran}
\bibliography{reference}

\end{document}